\documentclass{article}

\usepackage{PRIMEarxiv}

\usepackage[utf8]{inputenc} 
\usepackage[T1]{fontenc}    
\usepackage{hyperref}       
\usepackage{url}            
\usepackage{booktabs}       
\usepackage{amsfonts}       
\usepackage{nicefrac}       
\usepackage{microtype}      
\usepackage{lipsum}
\usepackage{fancyhdr}       
\usepackage{graphicx}       
\graphicspath{{media/}}     
\usepackage{paralist}  
\usepackage{subcaption}

\usepackage{xcolor}
\usepackage{amssymb}
\usepackage{pifont}

\title{Evaluation of GPT-4V and Gemini in Online VQA} 

\author{
  Mengchen Liu* \\
  Microsoft \\
  \texttt{mengcliu@microsoft.com} \\
  \And
  Chongyan Chen\thanks{equal contribution} \\
  University of Texas at Austin \\
  \texttt{chongyanchen\_hci@utexas.edu} \\
  \And
  Danna Gurari \\
  University of Colorado Boulder\\
  \texttt{Danna.Gurari@colorado.edu}\\
}

\begin{document}
\maketitle
\begin{abstract}
While there is much excitement about the potential of large multimodal models (LMM), a comprehensive evaluation is critical to establish their true capabilities and limitations.  In support of this aim, we evaluate two state-of-the-art LMMs, GPT-4V and Gemini, on a new visual question answering dataset sourced from an authentic online question answering community.  We conduct fine-grained analysis by generating seven types of metadata for nearly 2,000 visual questions, such as image type and the required image processing capabilities.  Our zero-shot performance analysis highlights the types of questions that are most challenging for both models, including questions related to "puzzling" topic, with "Identification" user intention, with "Sheet Music" image type, or labeled as "hard" by GPT-4.

\end{abstract}


\section{Introduction}



Artificial general intelligence (AGI) stands as a paramount objective for many AI communities. For example, OpenAI articulates its mission as ``ensure that artificial general intelligence - AI systems that are generally smarter than humans - benefits all of humanity.''~\cite{openai2023agi}. Recent advances in large multimodal models (LMMs), such as GPT-4V~\cite{OpenAI2023GPT4TR} and Gemini~\cite{gemini2023google}, have ignited constructive discussions on which level of AGI we have achieved. Comprehensive evaluations of such LMMs are necessary to address this inquiry.  Already, several papers report GPT-4V's performance across diverse domains, such as medical image, autonomous driving and across diverse tasks, including image captioning, visual grounding, and visual question answering (VQA)~\cite{li2023comprehensive,wen2023road,yang2023dawn,li2023medical}.  In this paper, we contribute to this research trajectory by evaluating GPT-4V and Gemini for the visual question answering (VQA) task.

Among existing VQA datasets, we employ VQAonline~\cite{chen2023vqaonline} to evaluate GPT-4V and Gemini due to its alignment with the criteria for assessing the AGI level of an AI system~\cite{morris2023levels}.  Firstly, this dataset places a strong emphasis on ecological validity, i.e., aligning with real-world tasks.  Each question in the dataset originates from everyday online users, reflecting their authentic information needs, in contrast to previous VQA datasets where questions are often sourced from crowd-sourced workers or textbooks. Secondly, this dataset is effective in evaluating the generality of an LMM. The questions span over 100 topics and encompass 7 distinct user intentions.  Third, the dataset facilitates a fair comparison with human performance.  All answers within the dataset are contributed by authentic everyday users, and each ground truth answer is verified by the user who posed the question.

To comprehensively evaluate GPT-4V and Gemini on VQAonline, we selected a subset of nearly 2,000 visual questions and systematically extracted a diverse set of metadata for each one. This includes super-topics, user intentions, the required image processing capabilities to answer the visual question, types of images, the difficulty level of questions, and the required types of knowledge to answer the question. Leveraging this wealth of information, we then analyzed the \textbf{zero-shot} performance of GPT-4V and Gemini across these various metadata categories.  Our aim is to shed light on valuable future directions for developing LMMs by pinpointing shortcomings of some of today's most advanced LMMs.

\section{Dataset}
\label{sec:dataset}
We now briefly introduce the VQAonline dataset~\cite{chen2023vqaonline} and outline the process of selecting a subset of visual questions from it, which serves as the focus of our analysis.

\paragraph{Source.}
VQAonline originates from the online question-answering platform Stack Exchange. Each of the 64,696 samples within VQAonline consists of four integral components: a natural language question, a natural language context, an image, and a ground-truth answer.  These components were gathered from Stack Exchange users through four distinct user entry interactions.  The natural language question was acquired in the form of a post ``title,'' utilizing a text entry field paired with explicit instructions to ``be specific and imagine you’re asking a question to another person'' during the question composition. The natural language context was provided through a ``body'' text entry field that supported inline images.  Users were instructed to furnish ``all the information someone would need to answer your question.'' The image component was collected as a URL link embedded within the natural language context. Subsequently, these three pieces of information were made publicly available, allowing other users on the site to contribute answers. In the final step, the individuals who posted the visual questions had the authority to mark one of the answers as the ``accepted answer.'' This answer is treated as the ground-truth answer in VQAonline.

\paragraph{Subset selection.}  
Due to the high cost and rate limitations associated with calling GPT-4V and Gemini API, we select a subset of visual questions in VQAonline for our analysis.  Recognizing the uneven distribution of visual questions across various topics in VQAonline, we followed \cite{chen2023vqaonline} to randomly sample 20 visual questions from each topic.  In cases where a topic had fewer than 20 samples, we included all available VQAs in this topic.  This approach yields a curated subset comprising \textbf{1,903} samples.

\section{Evaluation Methodology}
\label{section:evaluation-methodology}
Our goal is to comprehensively evaluate GPT-4V and Gemini on an authentic end-to-end VQA task. To this end, we first obtain the \textbf{zero-shot} predicted answer of GPT-4V and Gemini for each visual question.  Then, we adopt an evaluation methodology proven to align well with human expert judgments in VQAOnline~\cite{chen2023vqaonline} of prompting GPT-4 to rate the predicted answer by comparing it to the ground-truth answer.  Our analysis spans seven factors based on a combination of authentic metadata, specifically topic type, as well as automatically generated metadata, specifically super-topic, user intention, image processing capabilities, image types, difficulty levels, and knowledge types.  The evaluation process is described further below:

\paragraph{Answer Generation.}
To capture the authentic behavior of GPT-4V and Gemini, we focus on the zero-shot setting without any prompt engineering techniques, such as chain-of-thoughts~\cite{wei2022chain}. We adopt the following straightforward prompt format: ``\{Image\} \{Question\} \{Context\}''.
For GPT-4V, we use GPT-4V Turbo from Azure OpenAI service~\cite{gpt4vturbo}. For Gemini, we use gemini-pro-vision API~\cite{geminiapi}.\footnote{Due to RAI filtering, GPT-4V refuses to generate 45 (2.3\%) answers, while Gemini refuses to generate 32 (1.7\%) answers.}

\paragraph{Correctness Generation.}
In alignment with the original VQAOnline work~\cite{chen2023vqaonline}, we utilize GPT-4 to assess answer correctness, driven by the following considerations: (1) Evaluating the correctness of predicted answers in VQAonline often requires expert knowledge, rendering manual assessment impractical and expensive; (2) Traditional VQA metrics, such as VQA accuracy or n-gram-based evaluation metrics often prove insufficient for evaluating the long-form answers typical in VQAOnline; and (3) VQAonline research demonstrates that the GPT-4 metric highly aligns with human expert judgments.  To assess correctness, we prompt GPT-4 to ``compare the ground truth and prediction from AI models and give a correctness score for the prediction. The correctness score is 0 (totally wrong), 1 (partially correct), or 2 (totally correct).''  Subsequently, we rescale this GPT-4-based metric to range from 0 to 1.

\begin{table*}[ht!]
\centering
\begin{tabular}{p{4.5cm}p{11cm}}
\toprule
Attribute         & Example \\
\midrule
Question           & What is the scientific name of this plant that is possibly called ogre ears?        \\ \hline
Context            & I already did a reverse image search and came up with ``ogre ears'' as a name for this plant. However, this name is associated with other plants: Crassula argentea and Crassula ovata. What is the scientific name of this plant? Thanks \\ \hline
Image              & \includegraphics[width=5cm]{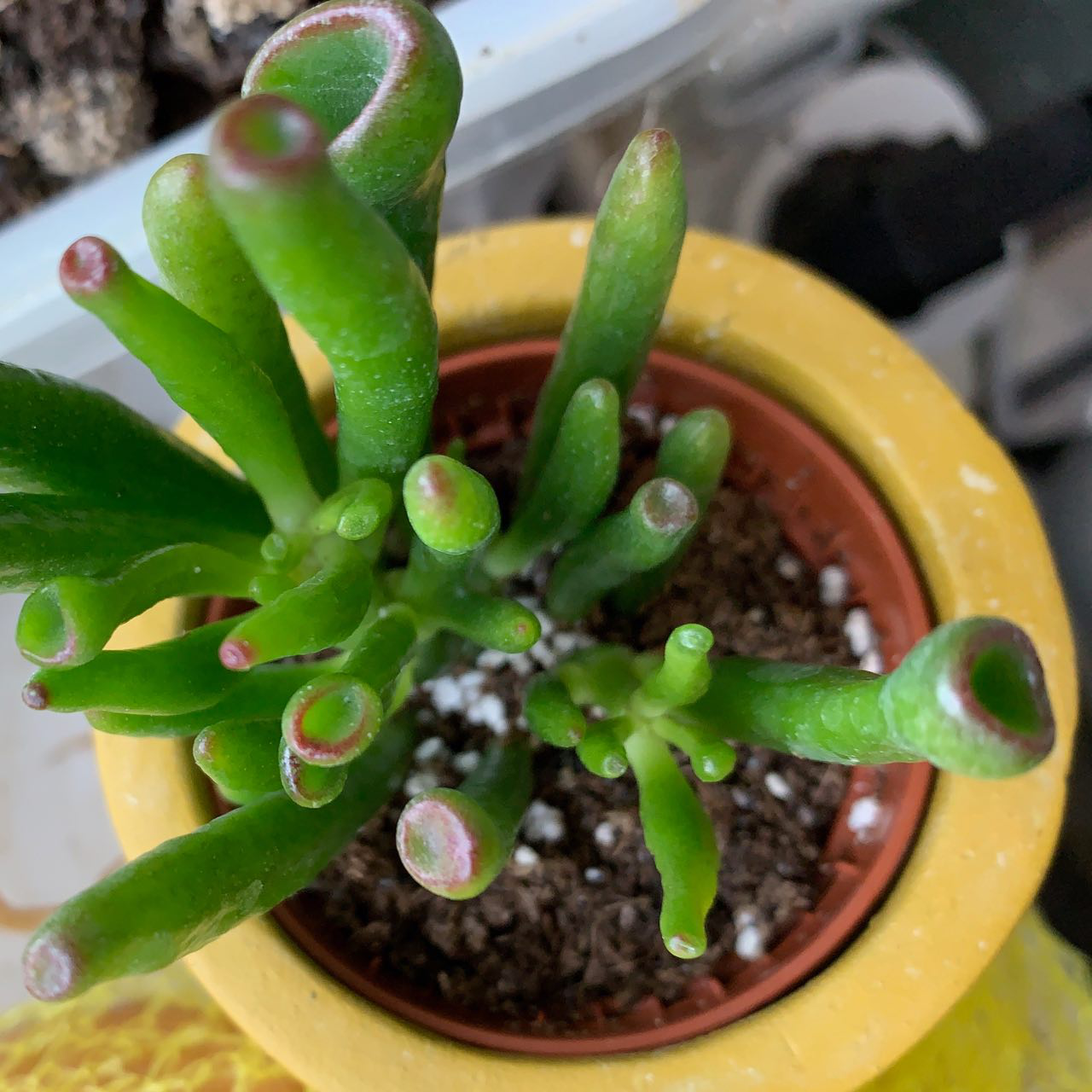} \\ \hline
Groundtruth Answer & This (lovely) succulent is Crassula ovata variety ``Gollum'', known also as ``gollum jade'' or ``finger jade''.  Search for the name and you'll find lots of commercial sources.  One informative source is here. \\ \hline
GPT-4V Answer      & The plant in the image is commonly referred to as ``Ogre Ears,'' but its scientific name is Crassula ovata `Gollum'. It is a cultivar of Crassula ovata and is characterized by its tubular, trumpet-shaped leaves that resemble the ears of the character Shrek or Gollum from ``The Lord of the Rings,'' hence the name. \\ \hline
Gemini Answer      & The scientific name of this plant is Crassula ovata 'Gollum'. It is a succulent native to South Africa. \\ \hline
Topic              & Biology \\ \hline
Super-topic        & Science and Nature \\ \hline
User Intention     & Identification \\ \hline
Image Processing Capabilities & Object Recognition \\ \hline
Type of Knowledge & Fine-grained World Knowledge \\ \hline
Image Type & Photographs \\ \hline
Difficulty Level & Medium \\ 

\bottomrule
\end{tabular}
\caption{A fully-coded sample in VQAonline to evaluate GPT-4V and Gemini.}
\label{table:evaluation-process}
\end{table*}
\paragraph{Metadata Generation.}
The VQAonline dataset provides a range of authentic metadata, including topics assigned to each sample in the dataset.  We augment this metadata leveraging GPT-4 and GPT-4V . A fully-coded sample is shown in Table~\ref{table:evaluation-process}.  In this example, the user queried the scientific name of a plant, accompanied by an image of that plant.

Different from previous VQA datasets, questions in VQAonline originate from genuine online users, reflecting their authentic information needs.  Given the diverse nature of information needs among everyday users, it is critical to comprehend the correlation between model performance and information needs, essentially, the underlying reasons prompting users to pose questions.  To unravel this connection, we utilize the following types of metadata:

\begin{itemize}
    \item \textbf{Topic}. This metadata is provided by the user who posed the question, reflecting the domain of the visual question, such as math or chess.  Topic samples and corresponding question samples can be found in Table~\ref{table:topic-sample} of Section~\ref{section:evaluation-results}. In total, our analyzed subset encompasses 106 topics spanning from everyday life to science.\footnote{We add one topic which is not included in the original VQAonline dataset: windows phone.}

    \item \textbf{Super-topic}. Given the large amount of topics, we summarize 10 super-topics from 106 topics with the assistance of GPT-4.  Specifically, we prompt GPT-4 to summarize super-topics using the topic names and their definitions. The resulting super-topics are shown in Table~\ref{table:super-topic} of Section~\ref{section:evaluation-results}.

    \item \textbf{User Intention}. Following the VQAonline dataset, we use 8 kinds of user intentions (Table~\ref{table:intention} in Section~\ref{section:evaluation-results}).  For each question, we prompt GPT-4 to generate the kind of user intention.
\end{itemize}

 
To correctly answer a visual question, a model must possess various capabilities,
including the ability to interpret and comprehend image content, a broad knowledge base, and the capacity to apply logical reasoning to information extracted from both the image and the accompanying text to formulate a coherent and precise answer. 
In this work, we focus on analyzing two specific types of capabilities, i.e., image processing capabilities and types of knowledge. This analysis enables a further examination of the specific capabilities where GPT-4V and Gemini excel or face challenges
\begin{itemize}

\item \textbf{Image Processing Capabilities}. 
    In contrast to textual QA, VQA differs as it requires visual information. In the VQAonline dataset~\cite{chen2023vqaonline}, over 99\% images are relevant to the question or context, with 66\% being essential for answering the question.  Thus, we investigate what kinds of image processing capabilities are needed to answer the visual questions.  We define 7 kinds of image processing capabilities as outlined in Table~\ref{table:capability} of Section~\ref{section:evaluation-results}.  Subsequently, for each visual question, we extract these essential capabilities to answer the question using GPT-4V.
    
    \item \textbf{Types of Knowledge}. 
    Building upon the work of Li et al.~\cite{li2023comprehensive}, we define 3 types of knowledge (Table~\ref{table:knowledge} in Section~\ref{section:evaluation-results}).  Different from this pioneering effort,  we introduce ``expert knowledge'' as a distinct category, recognizing that many visual questions in the VQAonline dataset necessitate domain-specific expertise.
\end{itemize}

In alignment with recent research efforts~\cite{yue2023mmmu}, we broaden our metadata coverage for each visual question to comprehensively evaluate GPT-4V and Gemini: 

\begin{itemize}
    \item\textbf{Difficulty Level}. To understand the performance difference on questions with different difficulty levels, we followed MMMU~\cite{yue2023mmmu} to categorize each visual question into four difficulty levels: 0 (very easy), 1 (easy), 2 (medium), and 3 (hard). For each visual question, we prompt GPT-4 to assign an appropriate difficulty level.

    \item \textbf{Image Type}. We observed that the visual questions in VQAonline have diverse image types. 
    Following the 30 types of image proposed in MMMU~\cite{yue2023mmmu}, we prompt GPT-4V to categorize the image in each visual question. A subset of image types and corresponding samples are illustrated in Figure~\ref{figure:image-type-sample} of Section~\ref{section:evaluation-results}.
\end{itemize}

\section{Evaluation Results}
\label{section:evaluation-results}
In this section, we report the evaluation results with respect to each type of metadata. Overall, the average accuracy of GPT-4V and Gemini is 0.53 and 0.42, respectively.

\subsection{Topics}
\label{section:topics}
Examining the model performance across various topics facilitates a comprehensive understanding of the suitable and challenging domains for an AI model. To date, the most related dataset is the MMMU dataset~\cite{yue2023mmmu} which contains 30 topics, primarily centered around subjects for college-level exams. We broaden this coverage, spanning science topics like math and physics to everyday-life-related subjects such as coffee and woodworking. Table~\ref{table:topic-sample} shows sample topics, their definitions (extracted of the website introduction), and example questions.

\begin{figure*}[htbp]
\centering  
\begin{minipage}{.33\textwidth}  
  \centering
  \includegraphics[width=1\linewidth]{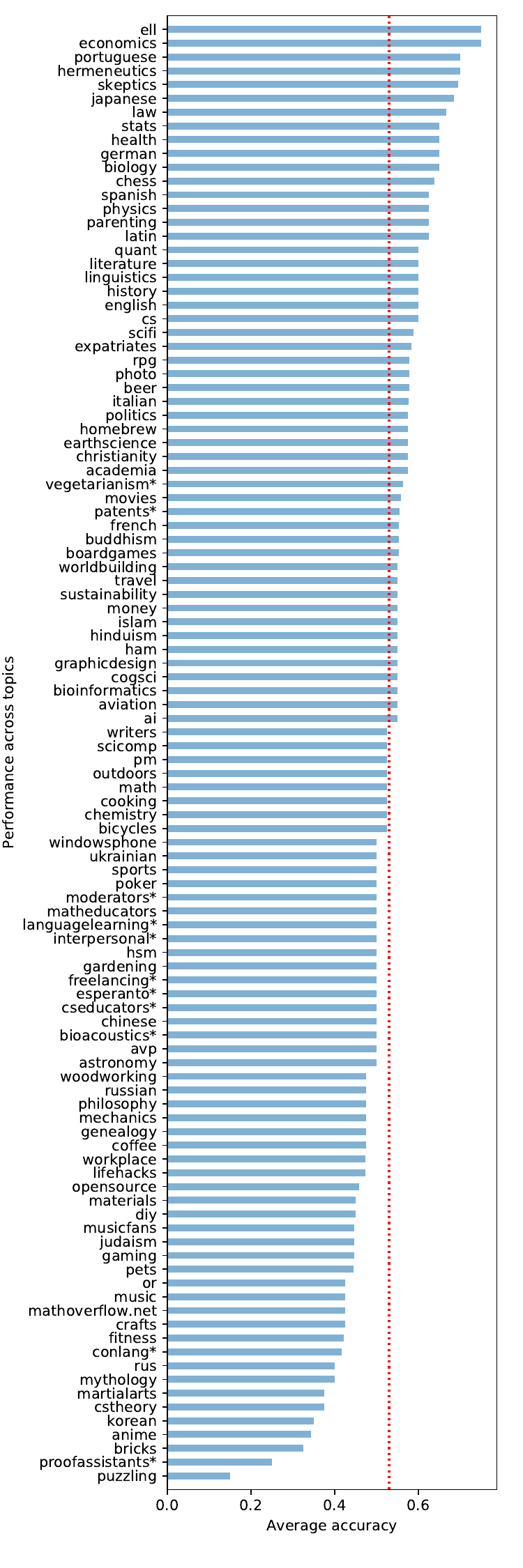}
  \captionof{figure}{GPT-4V Accuracy.}  
  \label{figure:topic-gptv}
\end{minipage}%
\begin{minipage}{.33\textwidth}  
  \centering
  \includegraphics[width=1\linewidth]{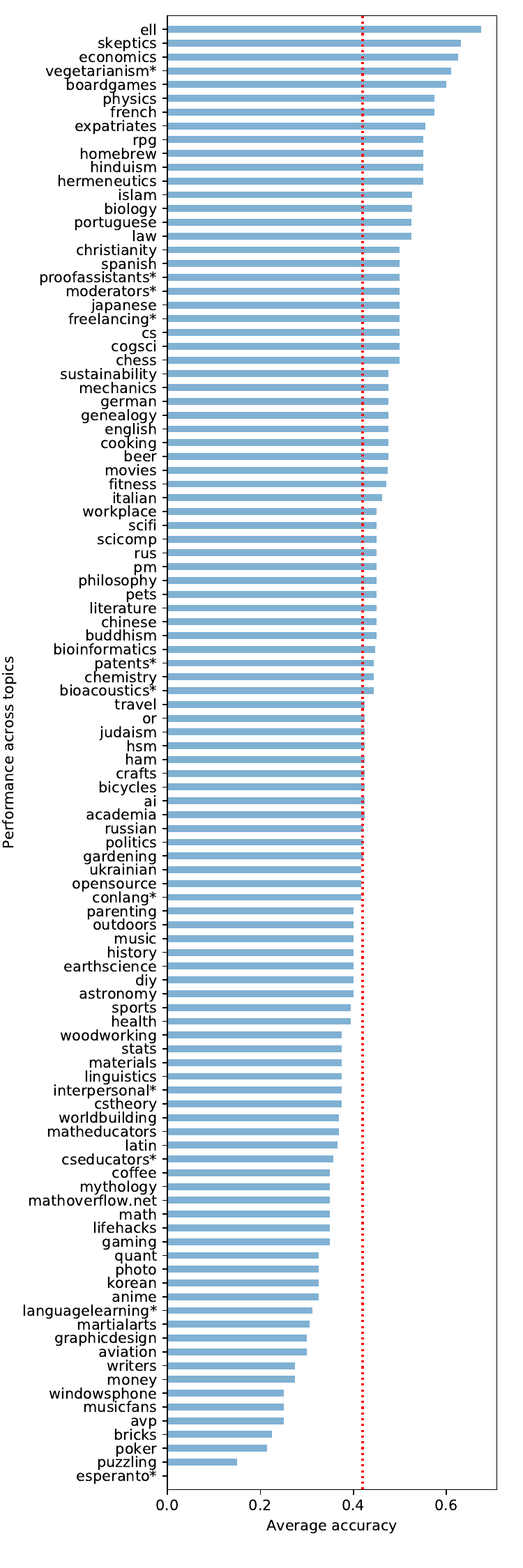}
  \captionof{figure}{Gemini Accuracy.}  
  \label{figure:topic-gemini}
\end{minipage}%
\begin{minipage}{.33\textwidth}  
  \centering
  \includegraphics[width=1\linewidth]{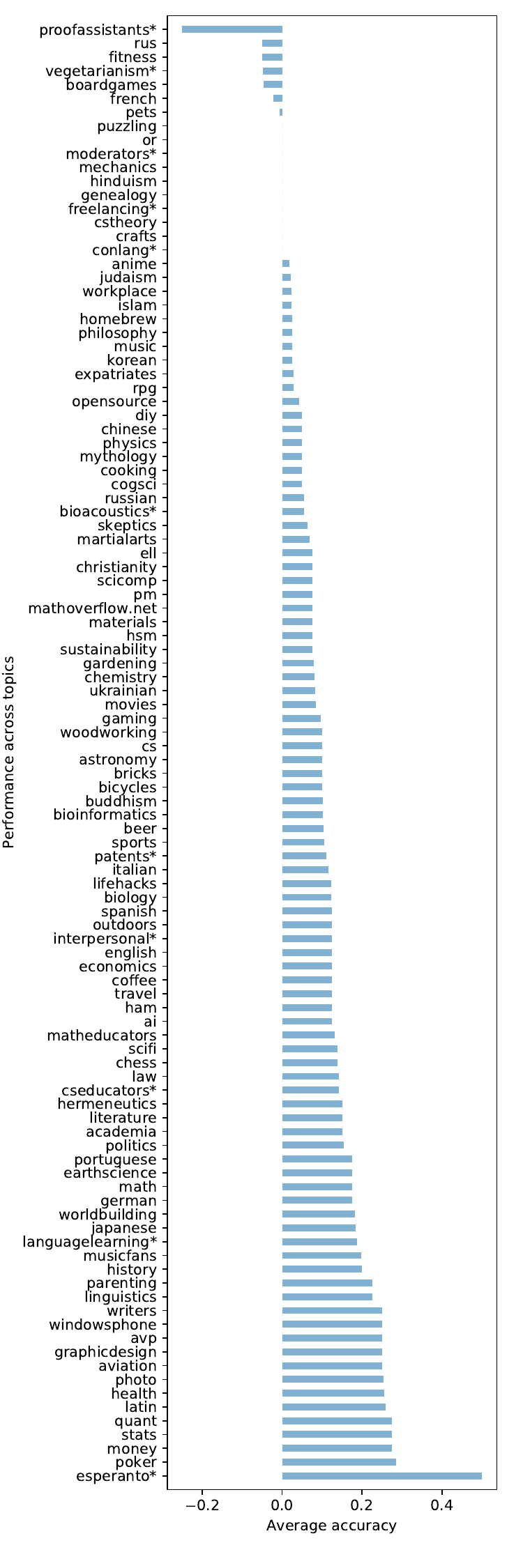}
  \captionof{figure}{Accuracy: GPT-4V - Gemini.}  
  \label{figure:topic-diff}
\end{minipage}%
\end{figure*}

The performance of GPT-4V and Gemini across 106 topics is illustrated in Figure~\ref{figure:topic-gptv} and \ref{figure:topic-gemini}, respectively. The average accuracy is denoted by a dotted line in the figures. Topics with fewer than 10 valid answers are marked with a $*$ in the figures.\footnote{For completeness, these topics are displayed in Figure~\ref{figure:topic-gptv},~\ref{figure:topic-gemini} and~\ref{figure:topic-diff}, but are excluded from the subsequent analysis.} 

Table~\ref{table:topic-best-worst} highlights the best- and worst-performing topics for GPT-V and Gemini. Topics where both models excel include ell (short for English language learners), economics, and skeptics (related to scientific skepticism). Conversely, topics posing challenges for both models are puzzling and bricks (related to LEGO). Example visual questions from these topics are presented in Figure~\ref{figure:puzzling},~\ref{figure:ell-1},~\ref{figure:economics-1}, and~\ref{figure:intention-identification-1}.

\begin{table*}[htbp]
\centering
\begin{tabular}{@{}p{1.7cm}p{5.8cm}p{8cm}@{}}
\toprule
 Topic     & Definition & Example questions \\ \midrule
 Biology   & ... for biology researchers, academics, and students.                              & What does "thermodynamic equilibrium" mean for an enzyme-substrate complex? \\ \hline
 Math      & ... for people studying math at any level and professionals in related fields. & How to calculate an unknown vector from a known vector and an angle? \\ \hline
 History   & ... for historians and history buffs.                                              & Was Robert McNamara present at Los Alamos, 1945?                                 \\ \hline
Woodworking & ... for professional and amateur woodworkers & How to securely attach a wooden plank to a wooden banister? \\

\bottomrule
\end{tabular}
\caption{Sample topics, their definitions, and question samples.}
\label{table:topic-sample}
\end{table*}

\begin{table*}[htbp]
\centering
\vspace{-10pt}
\begin{tabular}{@{}p{1.2cm}p{7.3cm}p{7cm}@{}}
\toprule
 & Best-5 Topics                                       & Worst-5 Topics                                   \\ \midrule
GPT-4V    & \textbf{ell}, \textbf{economics}, Portuguese, hermeneutics, \textbf{skeptics}       & \textbf{puzzling}, \textbf{bricks}, anime, Korean, cstheory \\
Gemini    & \textbf{ell}, \textbf{skeptics}, \textbf{economics}, boardgames, physics & \textbf{puzzling}, poker, \textbf{bricks}, avp,  musicfans   \\\bottomrule       
\end{tabular}
\caption{Best and worse-performing topics for GPT-4V and Gemini.}
\label{table:topic-best-worst}
\end{table*}
Figure~\ref{figure:topic-diff} illustrates the performance difference between GPT-4V and Gemini across topics, where the X-axis value represents the average accuracy of GPT-4V minus that of Gemini on each topic. Topics marked with $*$ have fewer than 10 valid answers in either model.
Across the majority of topics, GPT-4V performs better than Gemini.
\clearpage

\subsection{Super-topics}
\label{section:super-topics}
\begin{figure}[htbp]
     \centering
     \vspace{-20pt}
     \includegraphics[width=1\textwidth]{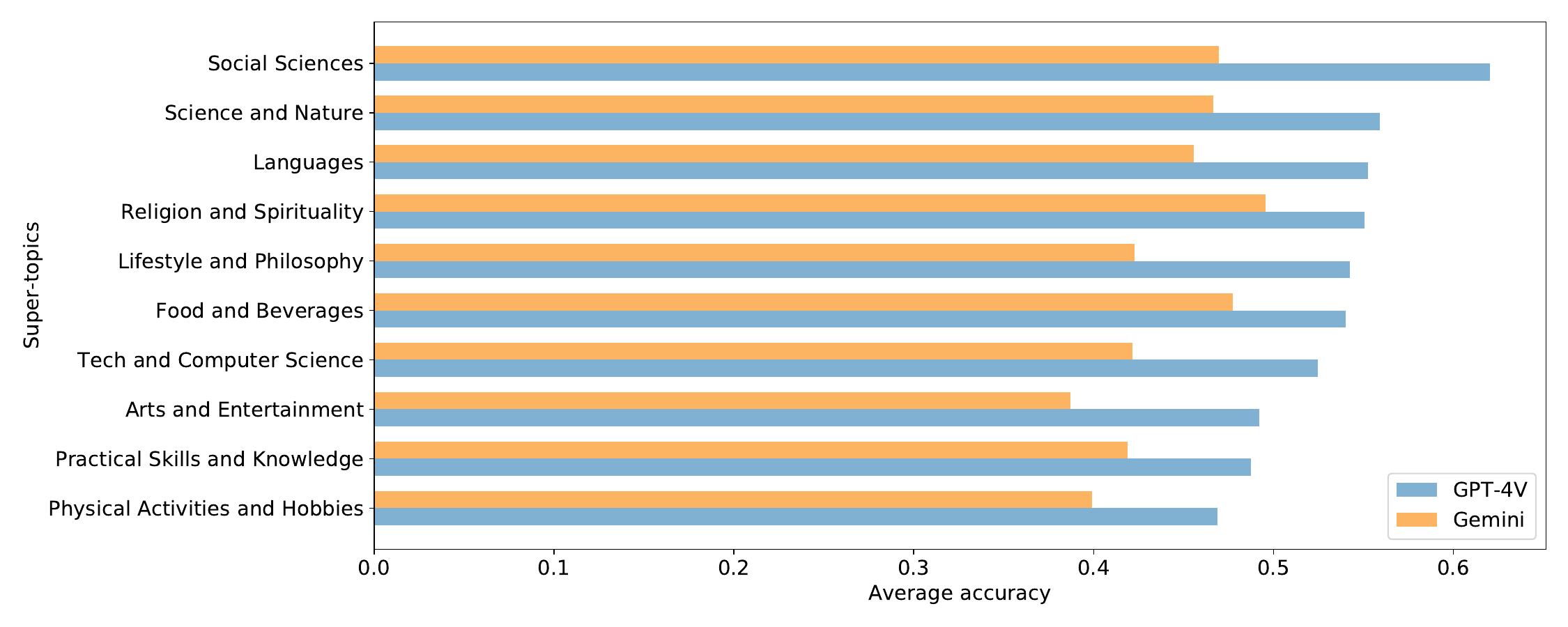}
     \vspace{-10pt}
        \caption{Accuracy comparison between GPT-4V and Gemini across super-topics.}
    \label{figure:super-topic}
\end{figure}

We then analyze the performances of GPT-4V and Gemini on super-topics to provide higher-level analysis.  For each super-topic in Table~\ref{table:super-topic}, we present the performance of GPT-4V and Gemini in Figure~\ref{figure:super-topic}. 

Notably, GPT-4V excels in science-related super-topics, particularly Social Sciences, followed by Science and Nature. However, it faces challenges in everyday-life-related super-topics such as Physical Activities and Hobbies. 

In contrast, Gemini exhibits its best performance in Religion and Spirituality, followed by Food and Beverages. It struggles most for the Arts and Entertainment super-topic. 

\begin{figure}[htbp]
     \centering
     \vspace{-15pt}
     \includegraphics[width=0.9\textwidth]{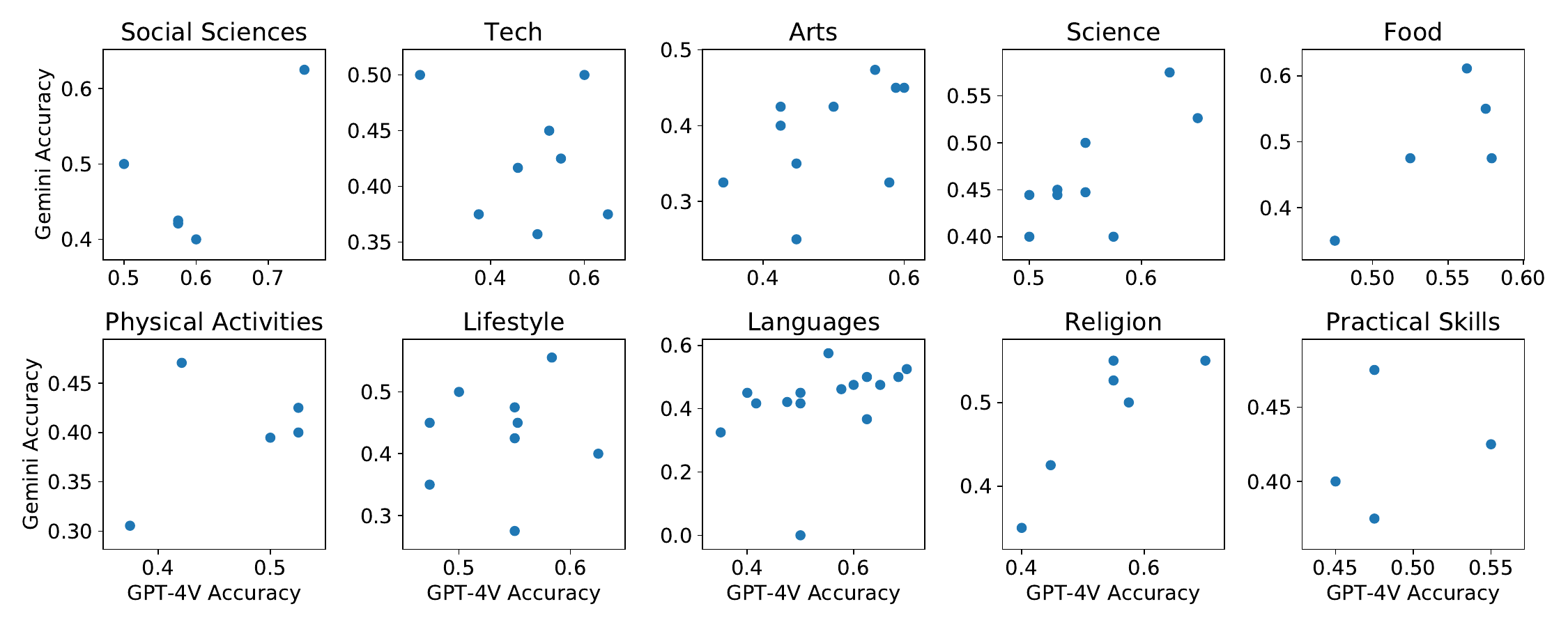}
     \vspace{-5pt}
        \caption{Correlation of topic-level accuracy between GPT-4V and Gemini. Each point is a topic, where its X-, and Y-coordinates are the accuracy in GPT-4V and Gemini, respectively.}
    \label{figure:super-topic-single}
\end{figure}

We further dive into each super-topic to study the correlation between GPT-4V and Gemini.  As illustrated in Figure~\ref{figure:super-topic-single}, on most super-topics, there is no visible correlation of accuracy between GPT-4V and Gemini. One exception is Languages, which shows a weak positive correlation ($R^2=0.34$). The outlier shown in Figure~\ref{figure:super-topic-single} is the topic Esperanto, for which we lack a sufficient number of valid answers from Gemini.

\vspace{-3pt}
\begin{table*}[htbp]
\centering
\begin{tabular}{p{4.5cm}p{11cm}}
\toprule
Super-topic                     & Topics                                                                                                                               \\ \midrule
Languages                       & conlang, japanese, esperanto, spanish, ukrainian, english, russian, latin, chinese, french, korean, italian, rus, german, portuguese \\\hline
Science and Nature              & cogsci, earthscience, biology, bioacoustics, physics, astronomy, chemistry, scicomp, bioinformatics                                  \\\hline
Food and Beverages              & vegetarianism, coffee, beer, cooking, homebrew                                                                                       \\\hline
Social Sciences                 & history, academia, politics, economics, moderators.                                                                                  \\\hline 
Arts and Entertainment          & musicfans, hsm, anime, crafts, movies, literature, music, photo, gaming, scifi                                                       \\\hline
Physical Activities and Hobbies & fitness, outdoors, sports, martialarts, bicycles                                                                                     \\\hline
Practical Skills and Knowledge  & woodworking, ham, mechanics, diy                                                                                                     \\\hline
Lifestyle and Philosophy        & buddhism, money, parenting, sustainability, lifehacks, travel, workplace, expatriates, freelancing                                   \\ \hline
Tech and Computer Science       & opensource, ai, proofassistants, stats, cseducators, cs, pm, cstheory                                                                \\\hline
Religion and Spirituality       & hermeneutics, islam, christianity, mythology, judaism, hinduism \\
\bottomrule
\end{tabular}

\caption{Super-topics and their corresponding topics.}
\label{table:super-topic}
\end{table*}

\clearpage
\clearpage
\vspace{-3pt}
\subsection{User intention}
\label{section:user-intention}

\begin{figure*}[htbp]

    \vspace{-5pt}
    \centering
    \begin{subfigure}{0.4\textwidth}
        \includegraphics[width=\linewidth]{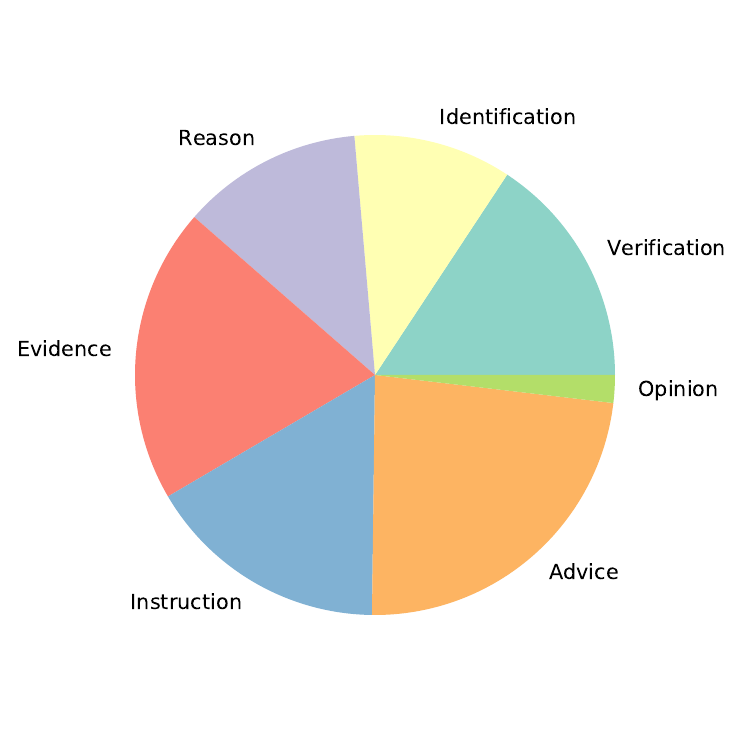}
        \caption{}
    \end{subfigure}
    \hfill  
    \begin{subfigure}{0.59\textwidth}
        \includegraphics[width=\linewidth]{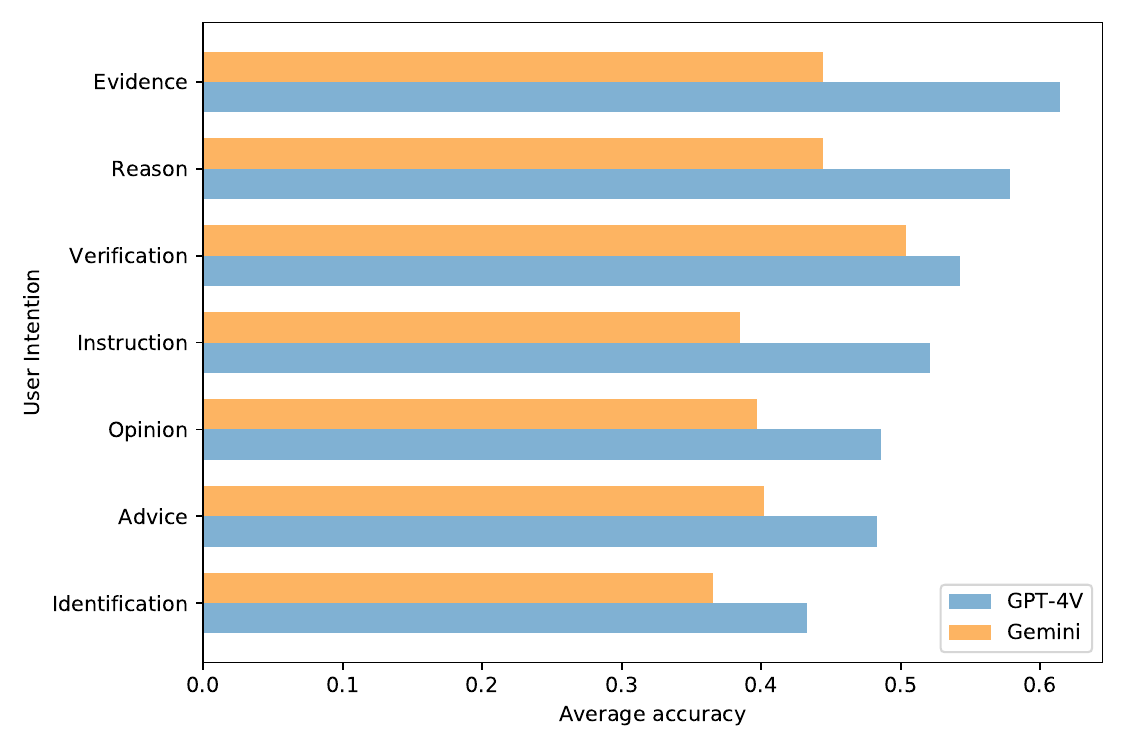}
        \caption{}
    \end{subfigure}
    \vspace{-5pt}
    \caption{User intention distribution and average accuracy: (1) user intention distribution in the subset of VQAonline; (b) accuracy comparison between GPT-4V and Gemini across user intentions.}
    \label{figure:intention}
\end{figure*}

Leveraging the 7 user intention categories outlined in VQAonline~\cite{chen2023vqaonline}, each visual question is assigned to one of the categories shown in Table~\ref{table:intention}, spanning common question types such as why, what, and how. The distribution of user intentions in our analyzed subset is illustrated in Figure~\ref{figure:intention}(a).  Except for ``Opinion'', which seeks for subjective opinions, other user intentions are well-balanced in our analyzed subset.

Figure~\ref{figure:intention}(b) illustrates the average accuracy comparison between GPT-4V and Gemini across various user intentions.
Notably, questions related to Identification pose the greatest challenge for both GPT-4V and Gemini. To delve deeper into this difficulty, we present several examples in Figure~\ref{figure:intention-identification-1} and~\ref{figure:intention-identification-2}.

\begin{table*}[htbp]
\centering
\begin{tabular}{p{2cm}p{13.5cm}}
\toprule
User intention                     &  Example question \\ \midrule
Verification & Did Cambridge award LLB? \\
\hline
Identification & What is this Lego piece - Yellow with handles and an octagonal section?\\
\hline
Reason & Why a blue spot in some hubble images of NGC 4302? \\
\hline
Evidence & What happens to jettisoned fuel tanks? \\
\hline
Instruction & How to replace background green color \\ 
\hline
Advice & What's a good (neutral) sample beer to spike with a sensory training kit? \\
\hline
Opinion  & Thoughts on today's article on ``farther'' vs ``further''? \\
\bottomrule
\end{tabular}
\caption{User intentions and example questions.}
\label{table:intention}
\end{table*}

    \vspace{-5pt}

\subsection{Image Processing Capabilities}
\label{section:capabilities}

We analyze the performance of GPT-4V and Gemini on questions that require different types of image processing capabilities. For this purpose, we prompt GPT-4 to generate the required image processing capabilities to answer a visual question, which is shown in Table~\ref{table:capability}.
Subsequently, for each question, we prompt GPT-4V to provide the necessary image processing capabilities.

\begin{figure*}[htbp]
    \centering
    \begin{subfigure}{0.6\textwidth}
        \includegraphics[width=\linewidth]{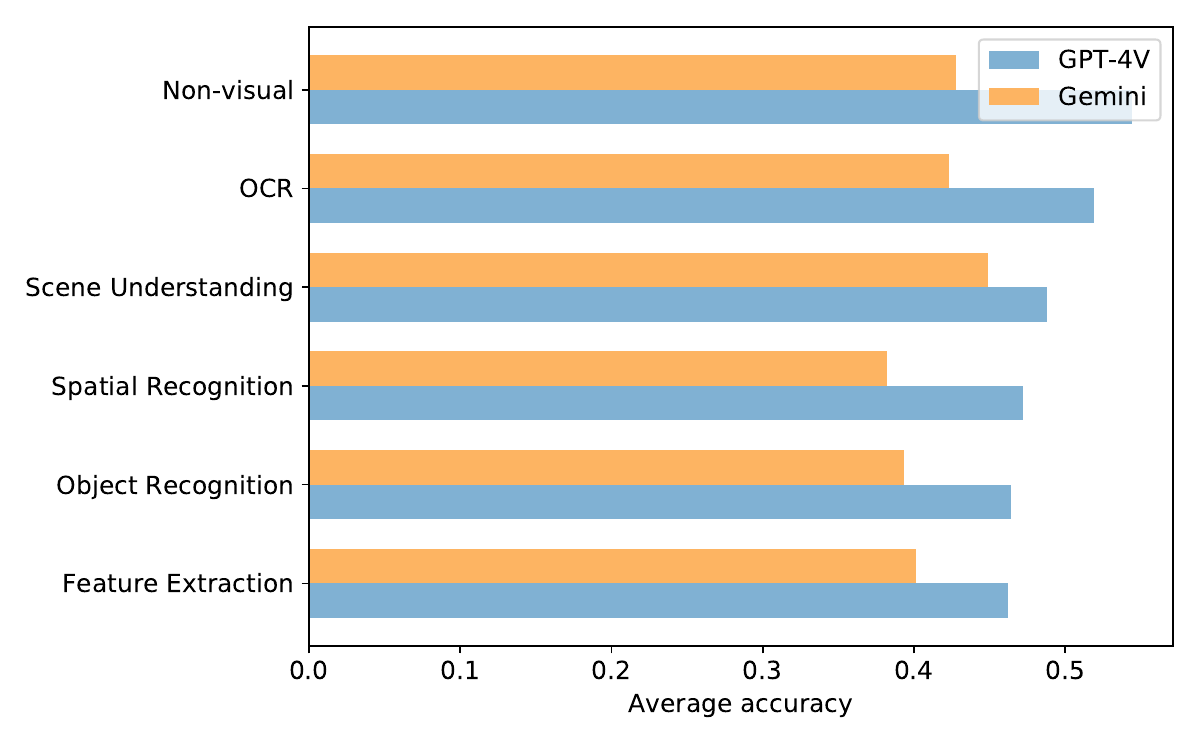}
        \caption{}
    \end{subfigure}
    \hfill  
    \begin{subfigure}{0.39\textwidth}
        \includegraphics[width=\linewidth]{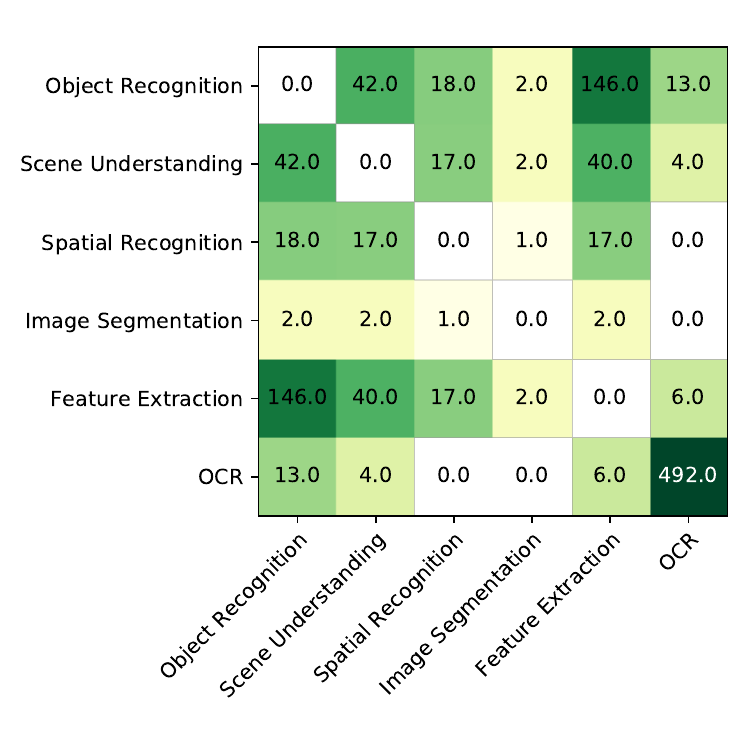}
        \caption{}
    \end{subfigure}
    \caption{Analysis of image processing capabilities: (a) Accuracy comparison between GPT-4V and Gemini on questions that required different image processing capabilities; (b) correlation between image processing capabilities.}
    \label{figure:capability}
\end{figure*}

As shown in Figure~\ref{figure:capability}(b), in our analyzed subset, visual questions related to OCR often exclusively require OCR capability.
On the other hand, feature extraction and object recognition are frequently simultaneously necessary. This underscores the importance for an LMM to possess multiple image processing capabilities concurrently, aligning with the recent advancements in foundation models. \footnote{We remove image segmentation and facial recognition due to too few valid samples in the subset and RAI-filtering of GPT-4V, respectively.}

The average accuracy across various image processing capabilities is shown in Figure~\ref{figure:capability}(a). GPT-4V excels particularly in non-visual questions, i.e., those requiring no image processing capabilities. Among the diverse image processing capabilities, GPT-4V finds ``Feature extraction'' to be the most challenging. Intriguingly, Gemini demonstrates its highest performance on questions related to ``Scene understanding''. This highlights distinctive strengths for Gemini in comprehending and responding to queries involving scene interpretation.
\vspace{-3pt}

\begin{table*}[htbp]
\centering
\begin{tabular}{@{}p{3cm}p{12.5cm}@{}}
\toprule
 Capability     & Definition \\ \midrule
Object Recognition & This is the ability to identify and classify different objects within an image. This is important for answering questions about what objects are present in the image. \\\hline
Scene Understanding & Beyond just recognizing individual objects, the AI needs to understand the scene as a whole. This might include understanding relationships between objects, or interpreting the overall context of the scene. \\\hline
Spatial Recognition & Understanding where objects are in relation to one another and to the scene as a whole is crucial. For example, a question might ask what is to the left of the tree in the image. \\\hline
Image Segmentation & Segmenting the image into different regions can help in identifying specific areas or objects referred to in the question. \\\hline
Feature Extraction & The AI needs to understand and extract relevant features from the image that could be important for answering the question. These could include things like color, texture, shape, size, and more. \\\hline
Optical Character Recognition (OCR) & If there are any textual elements in the image, OCR could be used to convert the text in the image into machine-readable text. \\\hline
Facial Recognition & If the question is about a person or people in the image, the AI might need the capability to recognize faces.\\ \bottomrule
\end{tabular}
\caption{Required image processing capabilities.}
\label{table:capability}
\end{table*}

\subsection{Knowledge Types}
\label{section:knowledge}
Following prior research~\cite{li2023comprehensive}, we categorize the required knowledge for answering each visual question into three distinct types: ``Everyday life knowledge'', ``Fine-grained world knowledge'', and ``Expert knowledge'', as illustrated in Table~\ref{table:knowledge}.
For each visual question, we prompt GPT-4 to generate the type of knowledge required to accurately answer the question. Instances where GPT-4 cannot determine the required knowledge type are marked as ``Other''.

\begin{figure*}[htbp]
    \centering
    \begin{minipage}{0.4\textwidth}
    
    \vspace{3pt}
        \includegraphics[width=\textwidth]{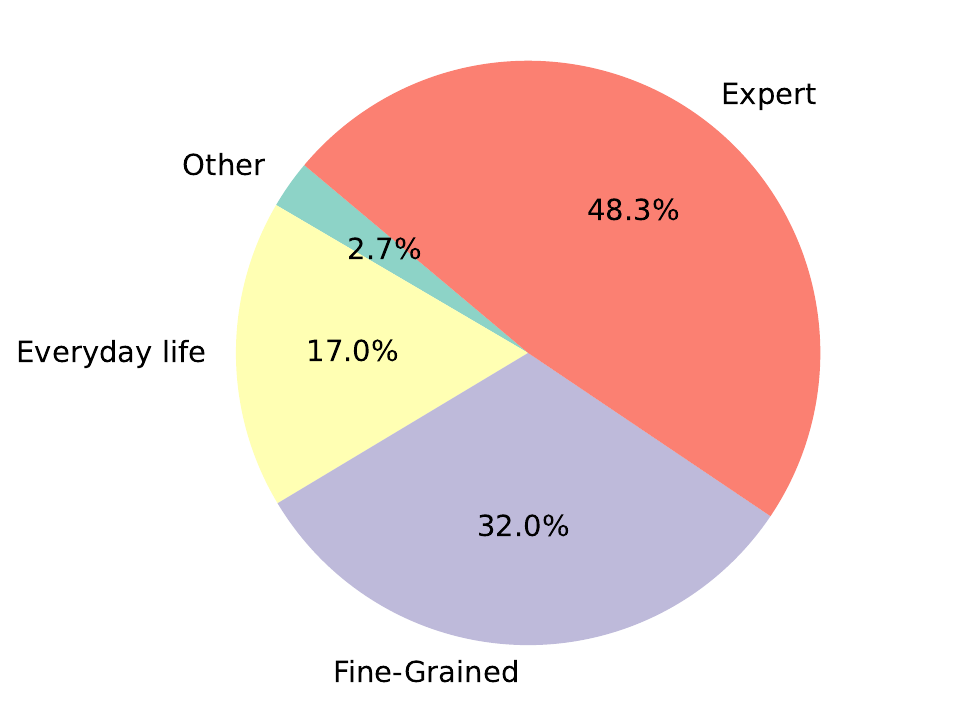}
    \vspace{-7pt}
        \caption{Distribution of knowledge types.}
        \label{figure:knowledge-distribution}
    \end{minipage}
    \hfill
    \begin{minipage}{0.59\textwidth}
        \includegraphics[width=\textwidth]{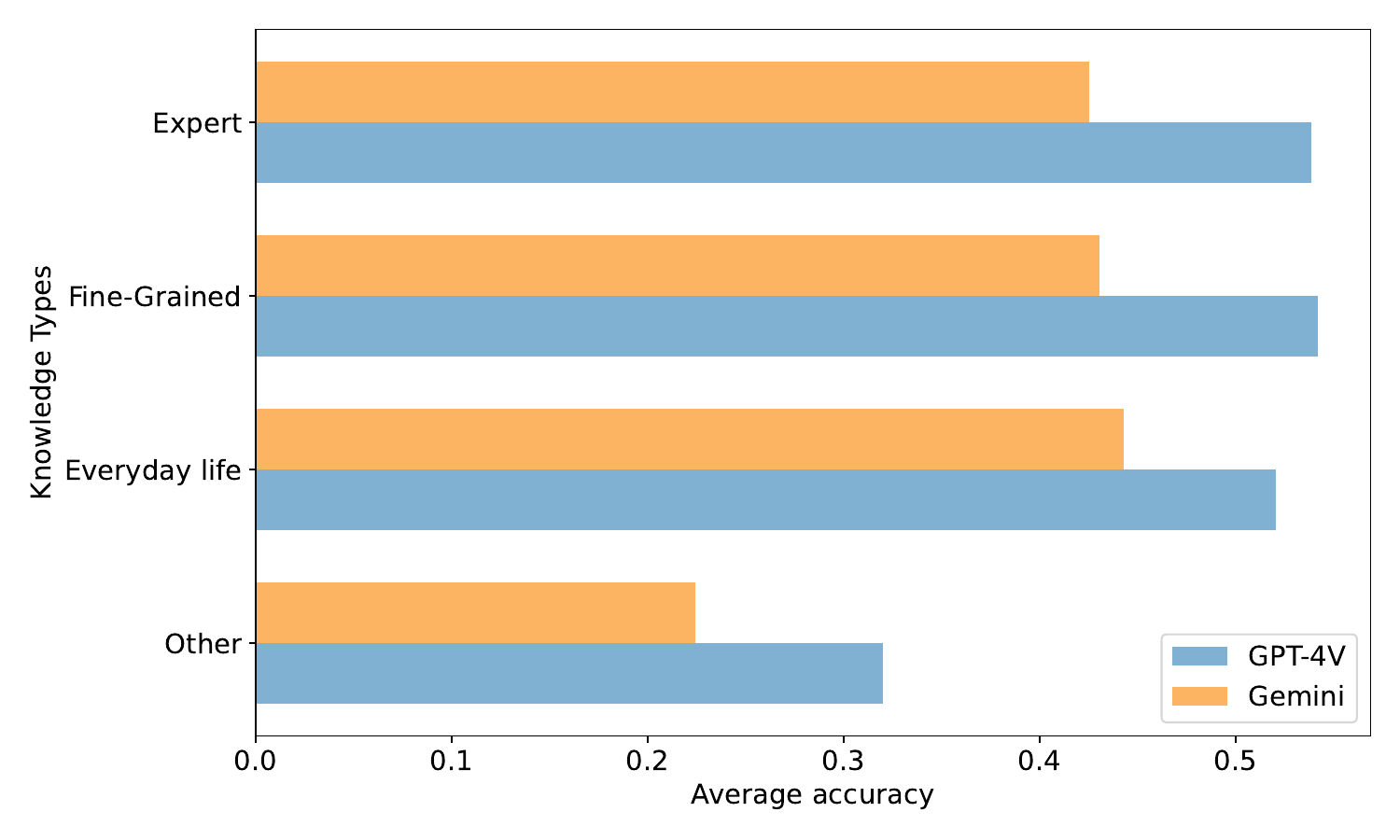}
        \caption{Accuracy comparison of GPT-4V and Gemini across different types of knowledge.}
        \label{figure:knowledge-accuracy}
    \end{minipage}
\end{figure*}

The distribution of knowledge types in our analyzed subset is presented in Figure~\ref{figure:knowledge-distribution}. Notably, the types of knowledge needed to correctly address visual questions in VQAonline exhibit diversity.  Approximately half of the visual questions require ``Expert knowledge,'' underscoring the significance of a model possessing domain-specific expertise to fulfill the diverse information needs of online users.

We compare the average accuracy between GPT-4V and Gemini in Figure~\ref{figure:knowledge-accuracy}. GPT-4V demonstrates a slightly better performance on questions requiring ``expert knowledge'' and ``fine-grained world knowledge'' compared to the other knowledge types. Conversely, Gemini performs slightly better on ``everyday life knowledge'' questions compared to the other knowledge types. 
Notably, both models exhibit their lowest performance on questions categorized as ``Other''.

\begin{table*}[htbp]
\centering
\begin{tabular}{p{3cm}p{6cm}p{3cm}p{3cm}}
\toprule
Knowledge Type                 & Definition & Example question &Example keywords \\ \midrule
Everyday life knowledge& It involves general knowledge about daily life and visual cues. It's broad, general and covers common, everyday experiences.& What is the use of this device? Perhaps a potato masher?&Safari, rabbit, balance ball, climbing shoes, soapy hands\\
\hline
Fine-grained world knowledge & It refers to specific information about the world which could include facts, information, and data about geography, history, science, current events, and more. This kind of knowledge is more detailed and specific. Such questions might be answered with web browsing. & Alternative source for instructions of LEGO set 7784 Batmobile?& I-9 form, Great Inflation in the 1970s/80s, heat blur from an aircraft engine
\\
\hline
Expert knowledge  & It is typically possessed by experts in a field, such as someone with a PhD.&How can I have a concerted addition of bromines? &Glycolysis, catalytic acid, Dictyostelium slug formation \\
\bottomrule
\end{tabular}
\caption{Knowledge types, their definition, example questions and example keywords.}
\label{table:knowledge}
\end{table*}

\subsection{Image Types}
\label{section:image-types}
To investigate the performance of models across the varied image types in VQAonline, we initially considered 30 categories, in alignment with the approach outlined in MMMU~\cite{yue2023mmmu}. Subsequently, we prompt GPT-4V to detect the specific image type in each visual question.
Among the 30 identified types, we focused our analysis on the 20 that had more than 10 instances. In addition, we also exclude the type ``Other''.  

\begin{figure*}[htbp]
     \centering
     \includegraphics[width=1\textwidth]{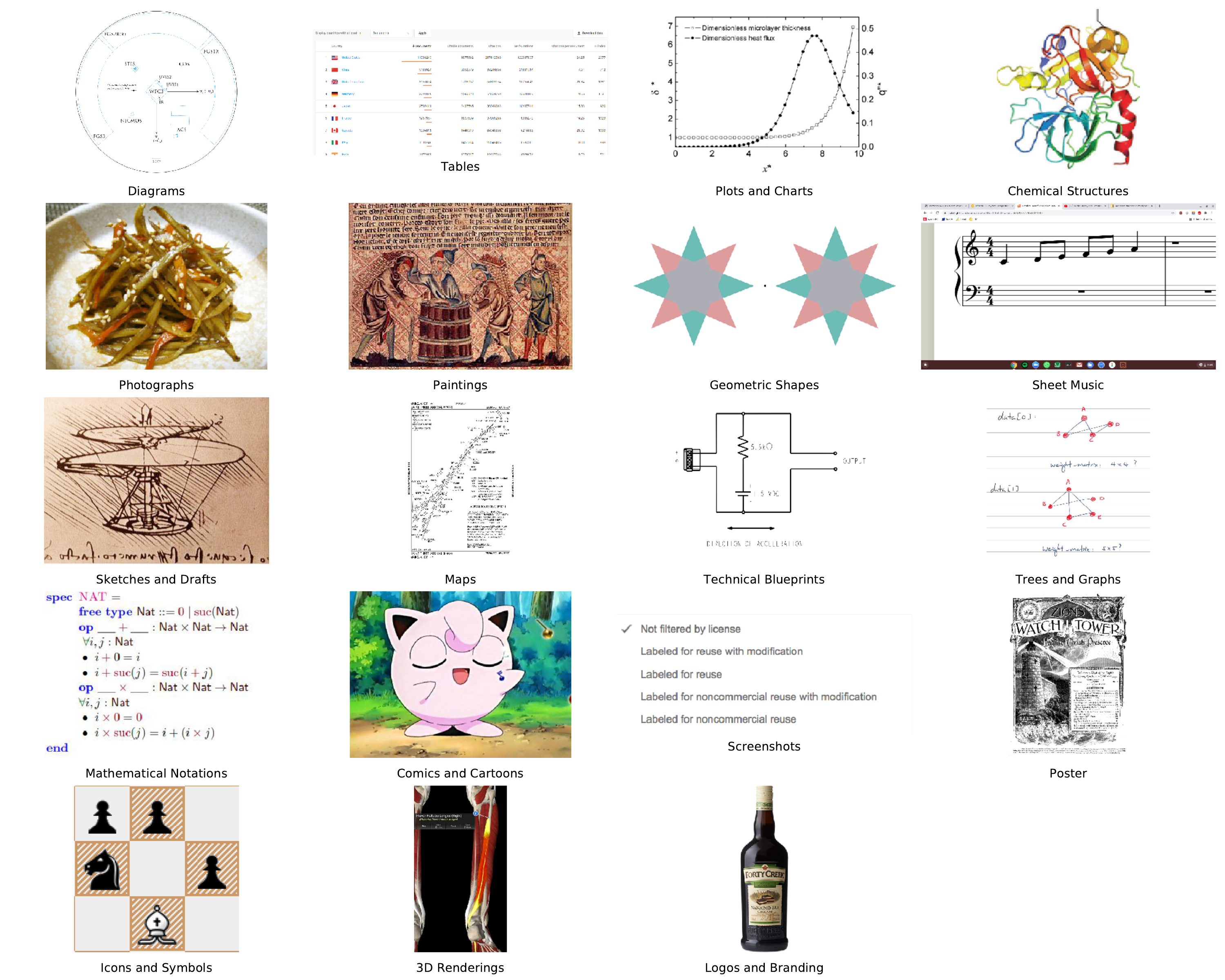}
        \caption{Sample for each image type.}
    \label{figure:image-type-sample}
\end{figure*}

A representative sample for each image type is presented in Figure~\ref{figure:image-type-sample}. Overall, we found that the Top-10 popular image types are Photographs (539), Screenshots (297), Plots and Charts(160), Tables (127), Mathematical Notations (77), Comics and Cartoons (75), Sketches and Drafts (71), Trees and Graphs (70),  Maps (54), and Diagrams (53). 

\begin{figure*}[htbp]
     \centering
     \includegraphics[width=0.7\textwidth]{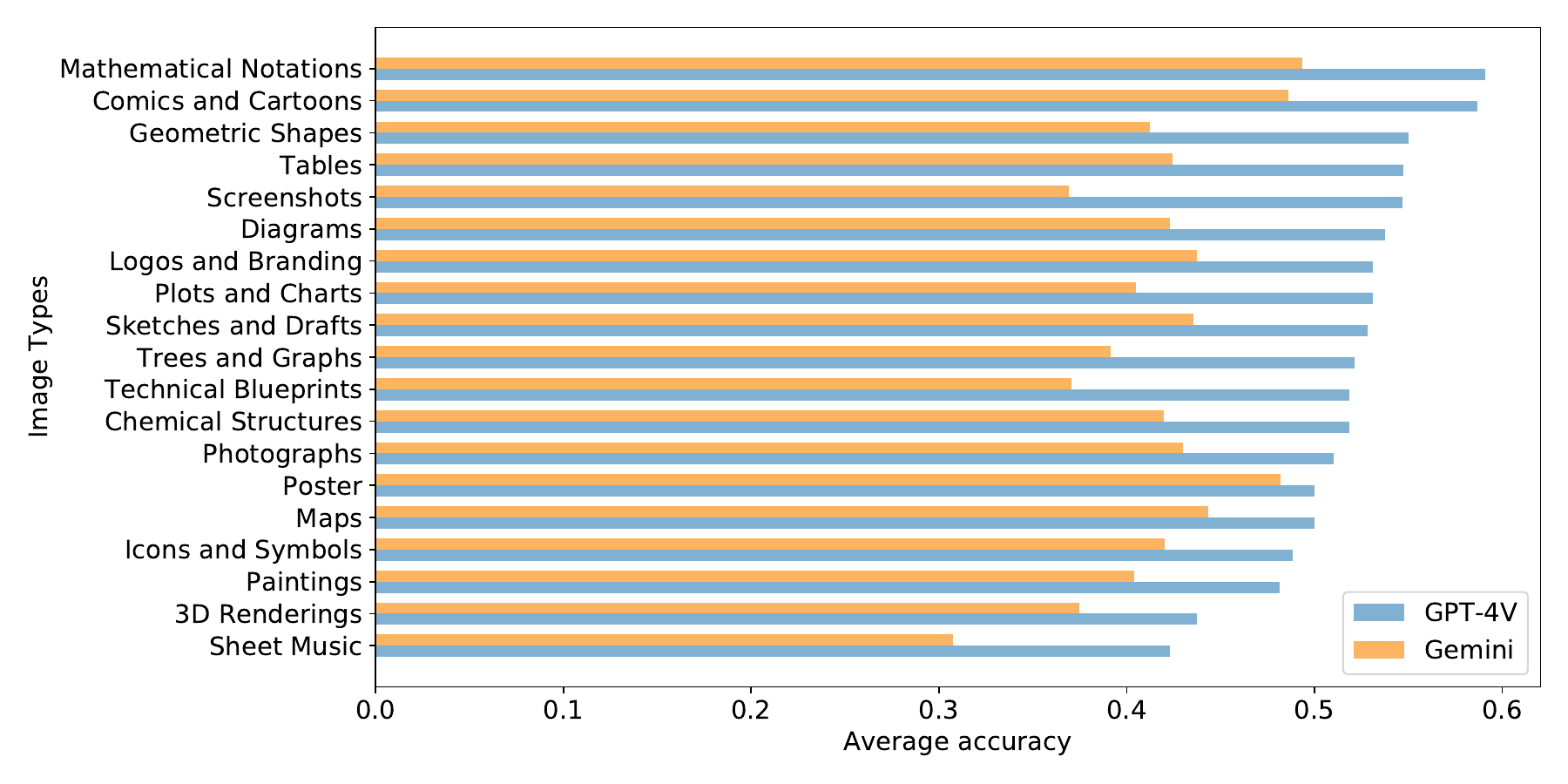}
        \caption{Accuracy comparison between GPT-4V and Gemini across image types.}
    \label{figure:image-type-accuracy}
\end{figure*}

Figure~\ref{figure:image-type-accuracy} illustrates the performance of GPT-4V and Gemini across image types. Notably, both models perform best on ``Mathematical Notations'' and ``Comics and Cartoons.'' Conversely,  both models perform worst on ``3D Renderings'' and ``Sheet Music.'' Interpreting the contents of these types of images often demands a robust foundation of domain-specific knowledge. An example visual question is illustrated in Figure~\ref{figure:music}.

\clearpage
\subsection{Difficulty Level}
\label{section:difficulty}
\begin{figure*}[htbp]
    \centering
    \begin{minipage}{0.4\textwidth}
    
    \vspace{3pt}
        \includegraphics[width=\textwidth]{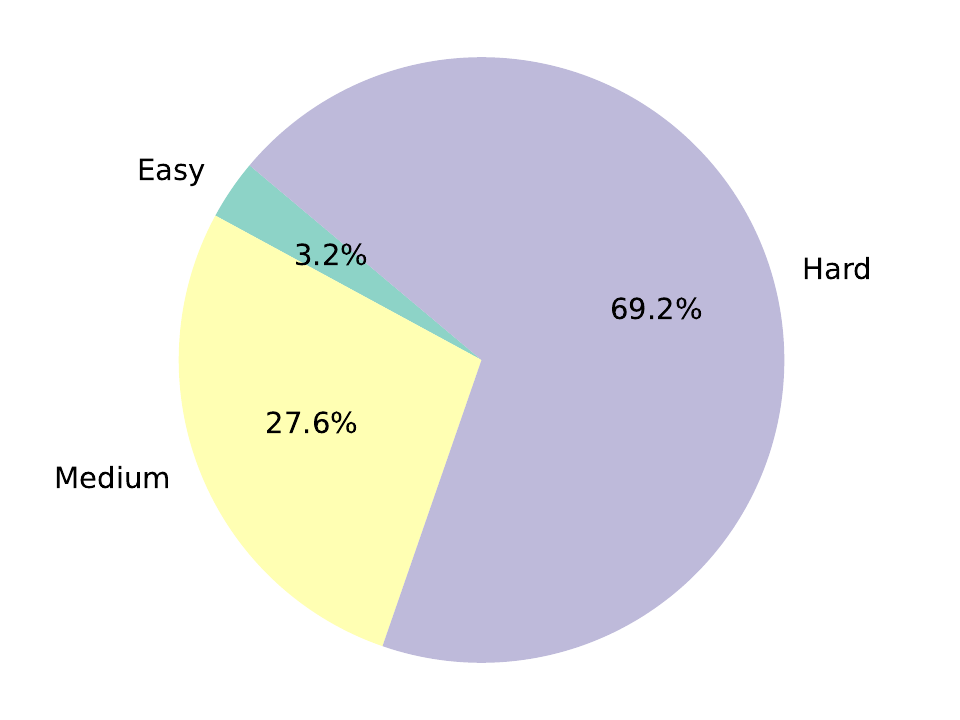}
    \vspace{-7pt}
        \caption{Distribution of question difficulty levels.}
        \label{figure:difficulty-distribution}
    \end{minipage}
    \hfill
    \begin{minipage}{0.59\textwidth}
        \includegraphics[width=\textwidth]{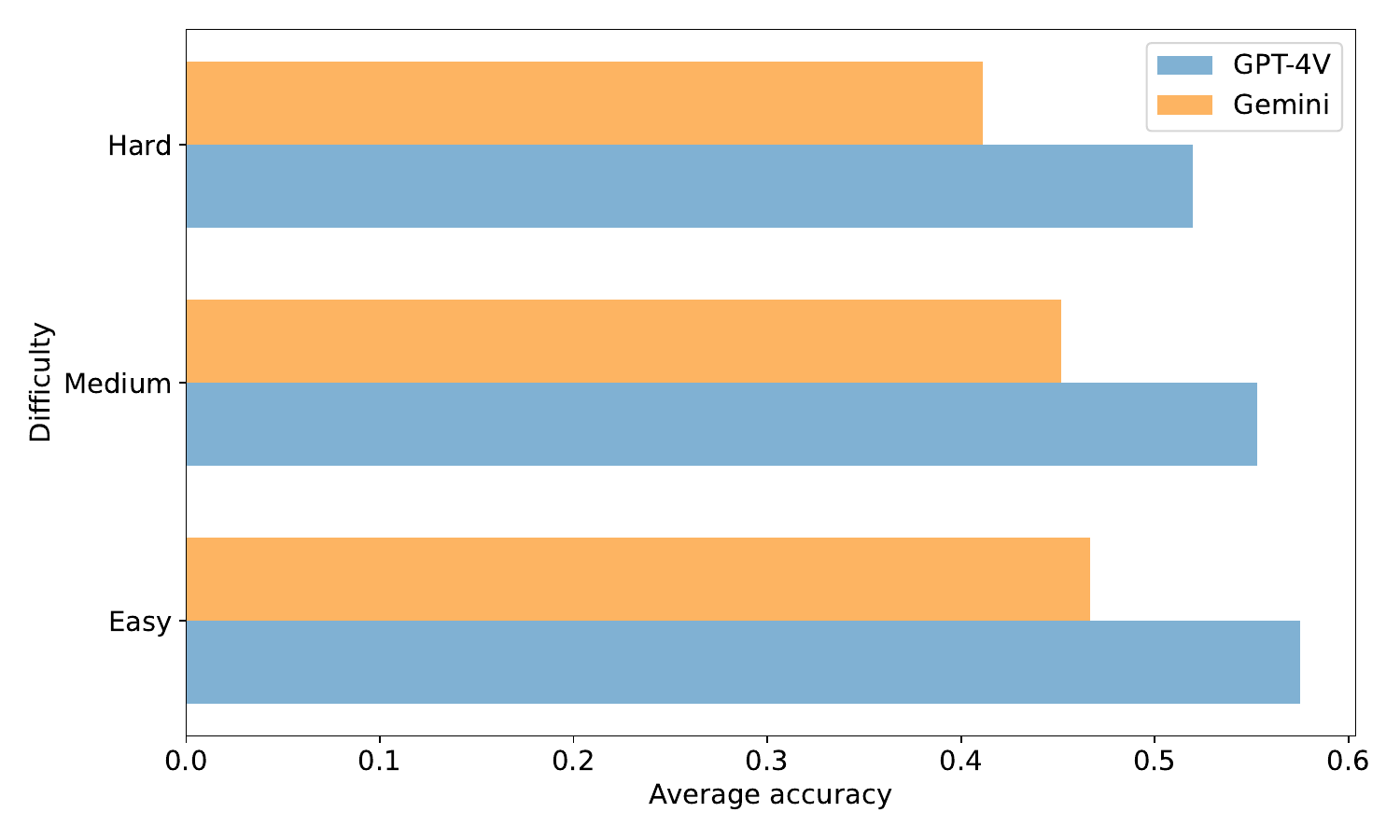}
        \caption{Accuracy comparison of GPT-4V and Gemini across different difficulty levels.}
        \label{figure:difficulty-accuracy}
    \end{minipage}
\end{figure*}

Question difficulty level is another dimension of variety in VQAonline. As shown in Figure~\ref{figure:difficulty-distribution}, around 2/3 of VQs are deemed hard. As expected, the accuracy of both models decreases for those harder questions (Figure~\ref{figure:difficulty-accuracy}).

\subsection{Take-away Notes}
We summarize the above analyses and discussions as follows:

\paragraph{Topics (Section~\ref{section:topics}).} 
Within the 106 topics covered in our analysis: GPT-4V and Gemini excel in ell (English language learners), economics, and skeptics (scientific skepticism). However, they encounter challenges in puzzling and bricks (LEGO-related).

\paragraph{Super-topics (Section~\ref{section:super-topics}).} 
Among the summarized 10 super-topics (Table~\ref{table:super-topic}): GPT-4V performs best on ``Social Sciences'' and ``Science and Nature'', 
while facing challenges in ``Physical Activities and Hobbies''.  Gemini performs best on ``Religion and Spirituality'' and worst on ``Arts and Entertainment''.

\paragraph{User intention (Section~\ref{section:user-intention}).}
Among the 7 user intentions identified in  Table~\ref{table:intention}, questions falling under ``Identification'' are the most challenging for GPT-4V and Gemini.

\paragraph{Image processing capabilities (Section~\ref{section:capabilities}).}
Among questions that require different image processing capabilities (Table~\ref{table:capability}), GPT-4V performs best on non-visual questions that don't require any specified capabilities, while Gemini demonstrates its highest performance on questions related to ``Scene understanding''. The most challenging visual questions for GPT-4V involve ``Feature extraction''.

\paragraph{Type of knowledge (Section~\ref{section:knowledge}).}
GPT-4V performs slightly better on questions demanding expert and fine-grained knowledge compared to those requiring everyday life knowledge; while Gemini performs slightly better on everyday life knowledge compared to the other types of knowledge.

\paragraph{Image types (Section~\ref{section:image-types}).}
In the spectrum of image types presented in Figure~\ref{figure:image-type-sample}, both GPT-4V and Gemini face most challenges with ``3D Renderings'' and ``Sheet Music''.

\paragraph{Difficulty level (Section~\ref{section:difficulty}).}
Across varying levels of question difficulty, GPT-4V and Gemini exhibit lower performance on more challenging questions in the VQAonline dataset.

\section{Related Works}
Existing assessments of GPT-4V and Gemini can be categorized into qualitative evaluations~\cite{wen2023road,yang2023dawn,li2023medical} and quantitative evaluations~\cite{yue2023mmmu,lu2023mathvista,yu2023mm}. It's important to note that these diverse efforts do not conflict but collectively contribute to a comprehensive evaluation of the state-of-the-art LMMs.

\paragraph{Qualitative evaluations.} 
These studies primarily rely on case-by-case analyses to assess the capabilities of GPT-4V and Gemini. For example, Yang et al.~\cite{yang2023dawn} explore GPT-4V with a collection of carefully curated qualitative samples spanning various domains and tasks. Insights derived from these samples showcase GPT-4V's ability to process arbitrarily interleaved multimodal inputs and the generality of its capabilities, positioning GPT-4V as a potent multimodal generalist system. While these evaluations provide intuitive insights into the strengths and weaknesses of the models, the absence of quantitative metrics makes it challenging to conduct fair comparisons between different models and with human performance.

\paragraph{Quantitative evaluations.}
Recently, several research efforts have provided a numerical evaluation of GPT-4V and Gemini. First-party evaluations from GPT-4V~\cite{OpenAI2023GPT4TR} and Gemini~\cite{gemini2023google} focus on analyzing the model performance across diverse tasks, such as coding, language understanding, and VQA. While these first-party evaluations provide a broad overview, a more comprehensive understanding of detailed model capabilities can be gained through in-depth analyses of specific tasks. Aligning with such needs, various recent works have undertaken a detailed examination of GPT-4V's performance in specific tasks~\cite{li2023comprehensive,lu2023mathvista,yue2023mmmu}. Among these efforts, the most closely aligned research to our evaluation is the analysis accompanying the MMMU dataset~\cite{yue2023mmmu}. The paper provides a thorough  analysis of GPT-4V in MMMU, which includes visual questions from college-level exams. In comparison, our evaluation focuses on comparing the performance of GPT-4V and Gemini. Furthermore, our evaluation better aligns with real-world scenarios, as all questions are sourced from authentic users on a popular, public community question answering platform. 

\section{Conclusion}
In this study, we present a thorough evaluation of GPT-4V and Gemini using the VQAonline dataset. For each visual question within our examined subset, we furnish a comprehensive set of metadata, encompassing topic, super-topic, user intention, image processing capabilities, image type, question difficulty, and the required type of knowledge. Our analysis delves into the performance disparities between GPT-4V and Gemini across these dimensions, pinpointing challenging questions for both models.

For future work, we intend to expand the analysis workflow to incorporate additional datasets. In addition, exploring the model performance when utilizing multiple types of metadata simultaneously presents an intriguing avenue for investigation.

\bibliographystyle{unsrt}  
\bibliography{references}  

\newpage
\appendix
\section{Additional Figures}

\begin{figure*}[h]
     \centering
     \includegraphics[width=1\textwidth]{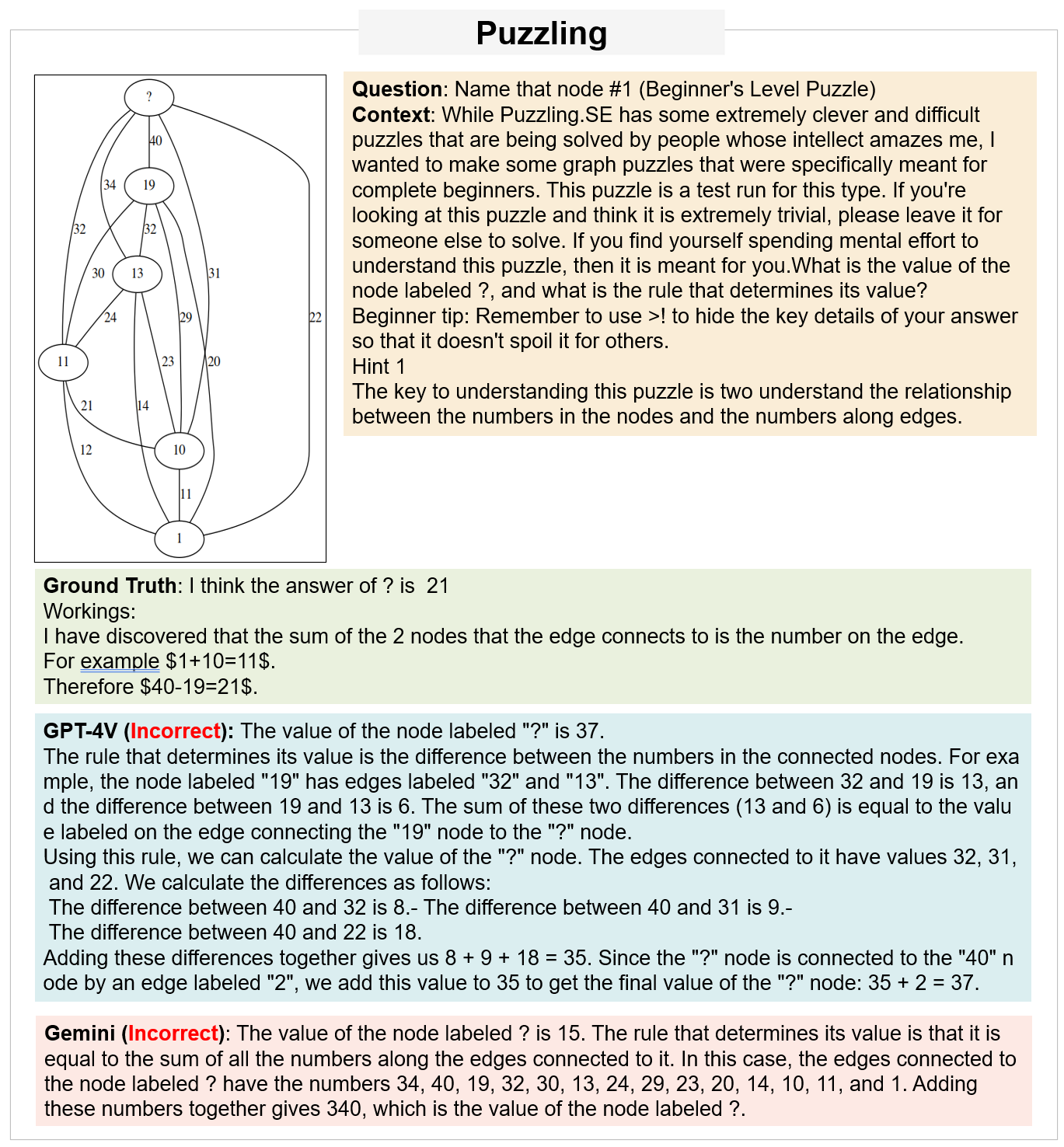}
        \caption{Sample from topic ``puzzling''.}
    \label{figure:puzzling}
\end{figure*}

\begin{figure*}[h]
     \centering
     \includegraphics[width=1\textwidth]{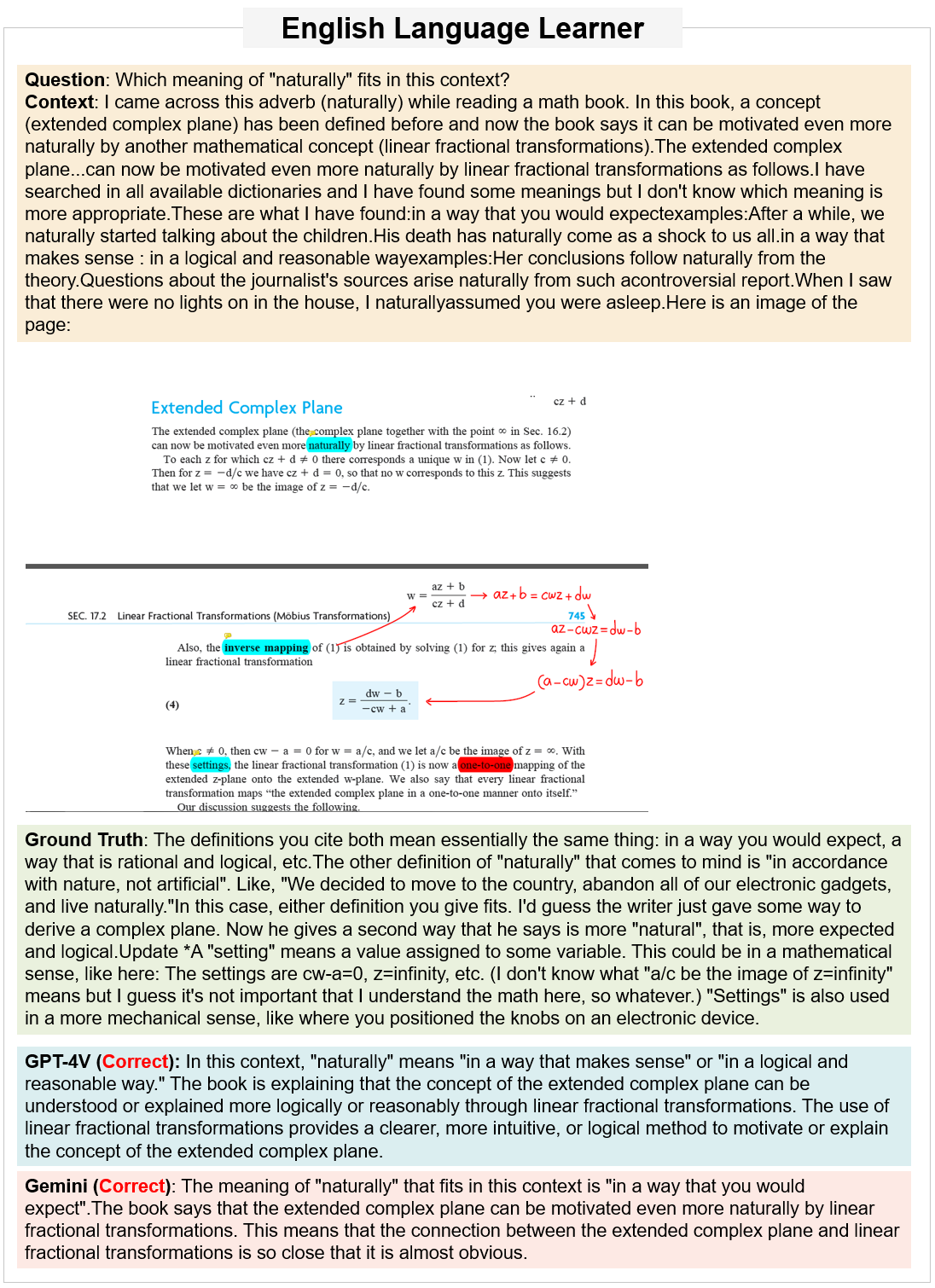}
        \caption{Sample from topic ``ell'' (English language learning).}
    \label{figure:ell-1}
\end{figure*}

\begin{figure*}[h]
     \centering
     \includegraphics[width=1\textwidth]{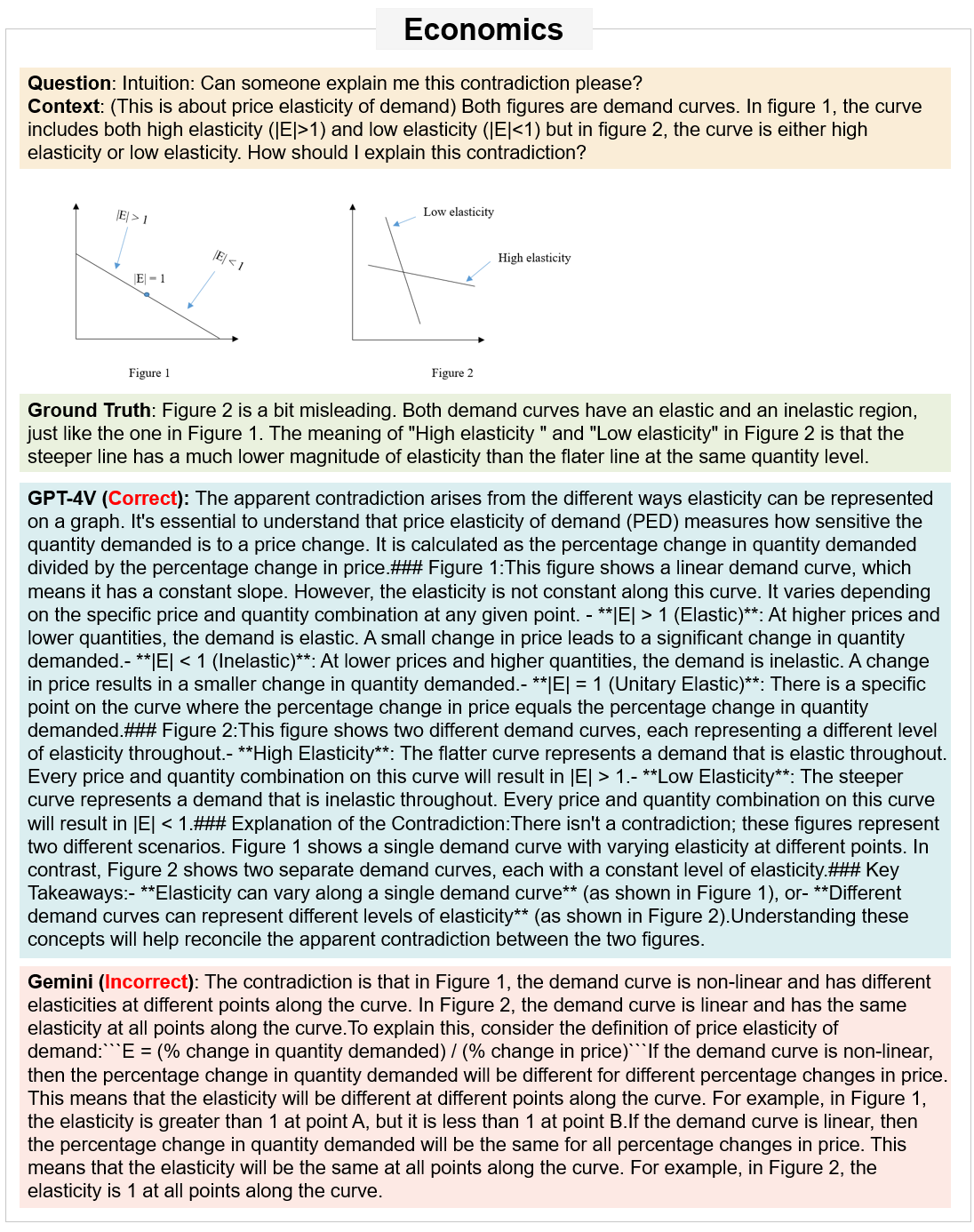}
        \caption{Sample from topic ``ecomonics''.}
    \label{figure:economics-1}
\end{figure*}

\begin{figure*}[h]
     \centering
     \includegraphics[width=1\textwidth]{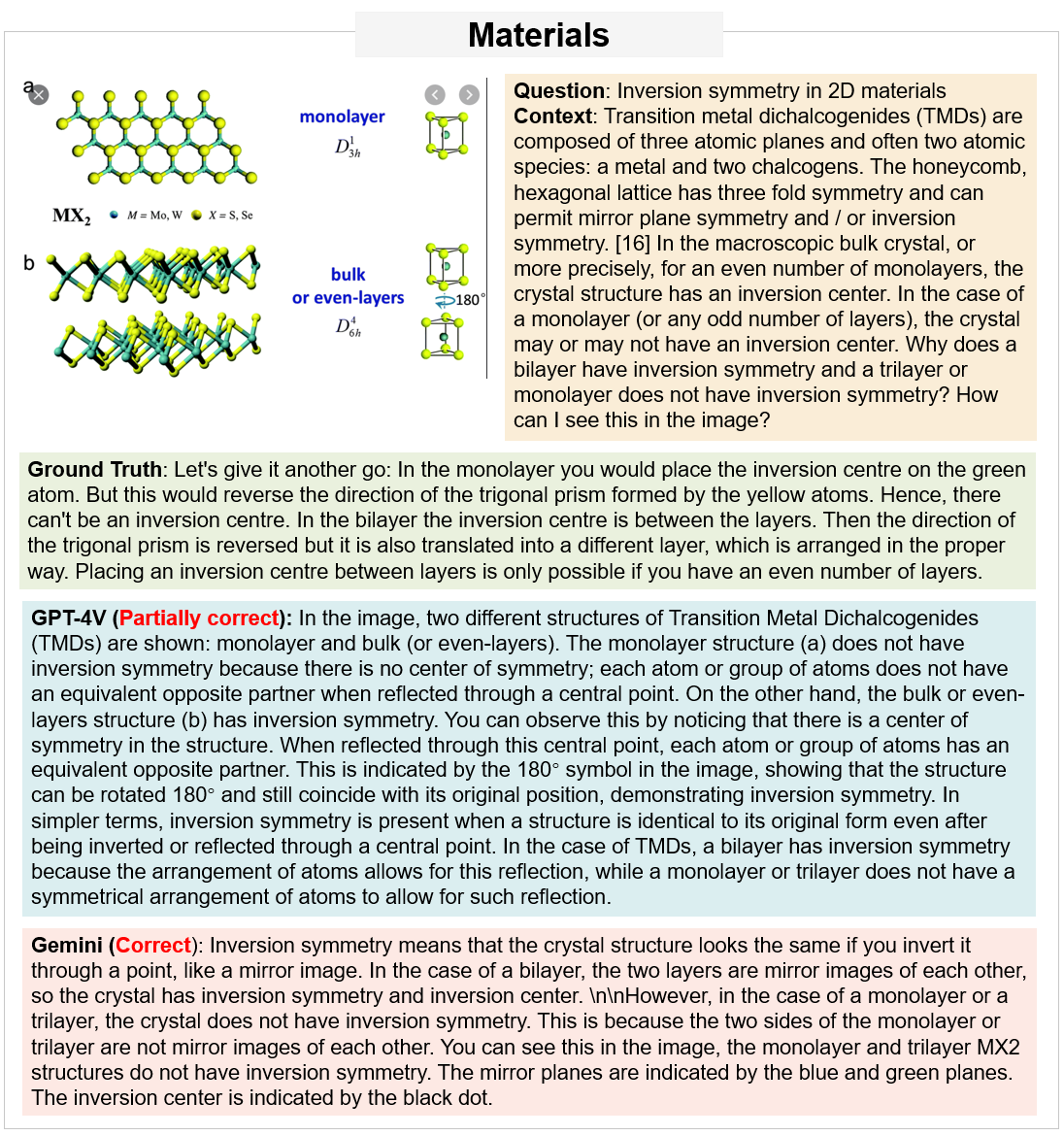}
        \caption{Sample from topic: materials.}
    \label{figure:materials}
\end{figure*}

\begin{figure*}[h]
     \centering
     \includegraphics[width=1\textwidth]{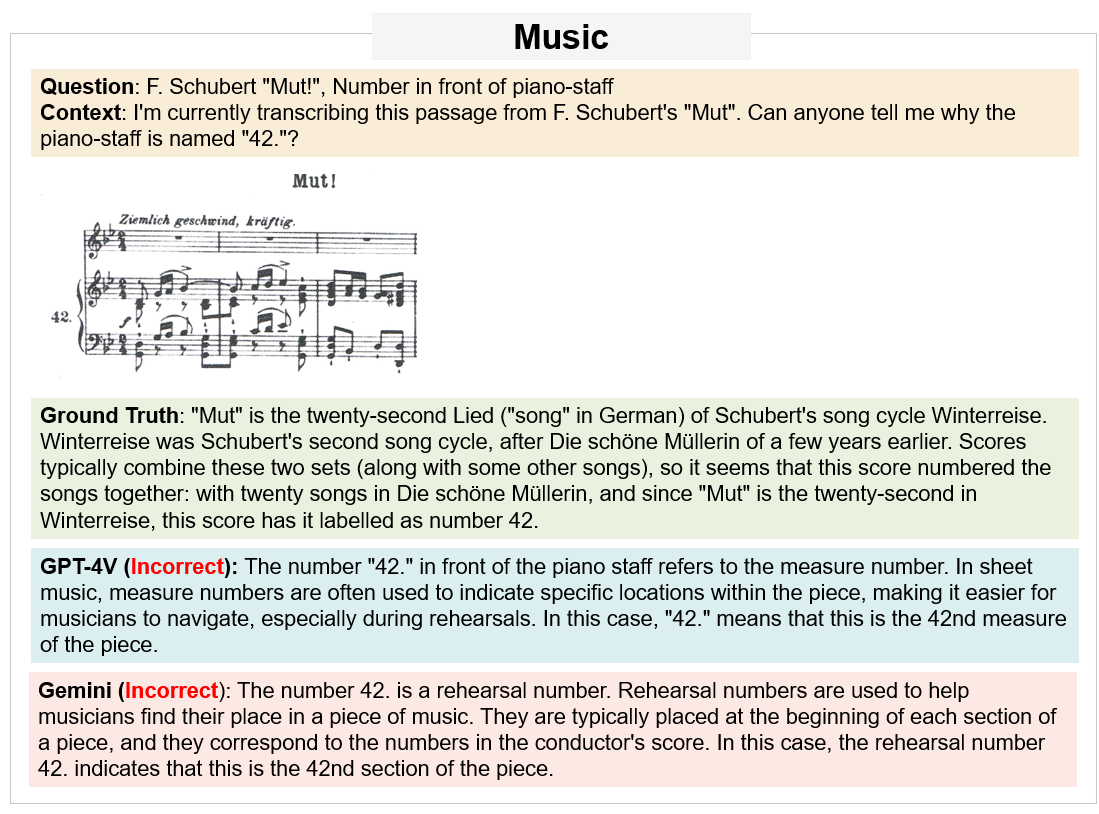}
        \caption{Sample from topic: music.}
    \label{figure:music}
\end{figure*}

\begin{figure*}[h]
     \centering
     \includegraphics[width=1\textwidth]{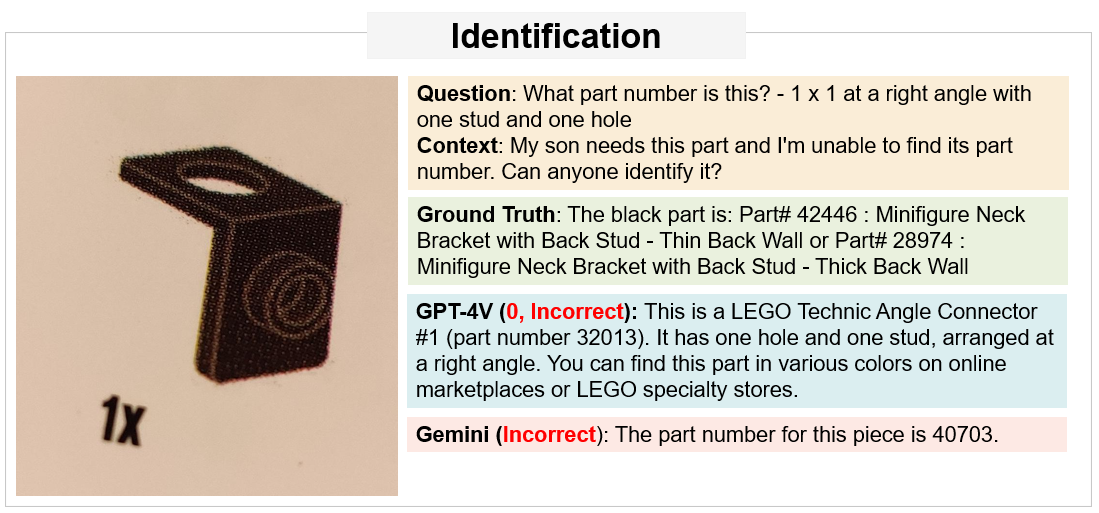}
        \caption{Sample of user intention: identification.}
    \label{figure:intention-identification-1}
\end{figure*}

\begin{figure*}[h]
     \centering
     \includegraphics[width=1\textwidth]{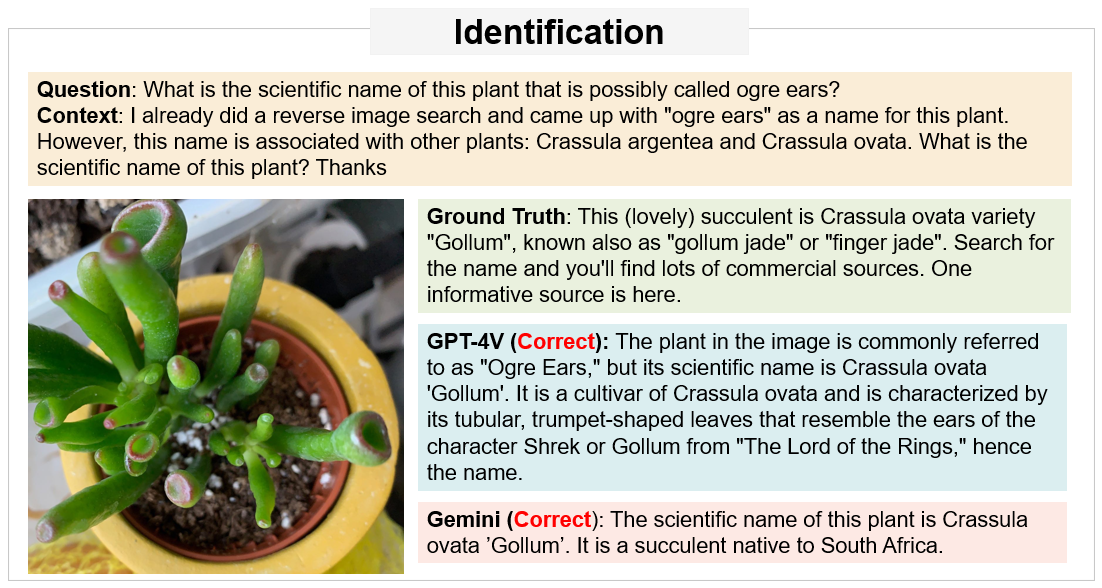}
        \caption{Sample of user intention: identification.}
    \label{figure:intention-identification-2}
\end{figure*}

\begin{figure*}[h]
     \centering
     \includegraphics[width=1\textwidth]{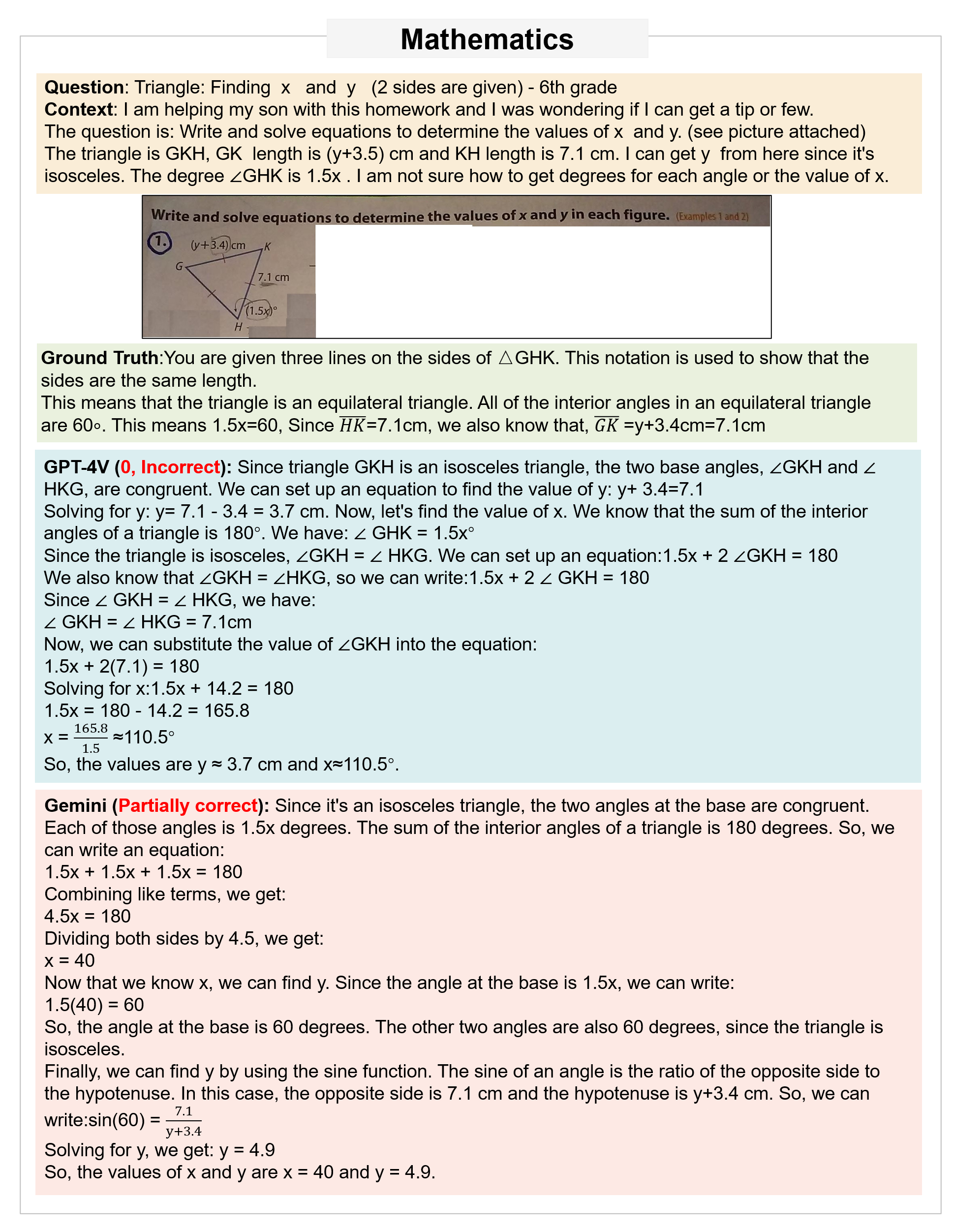}
        \caption{Sample of topic math.}
    \label{figure:math}
\end{figure*}

\end{document}